\documentclass[10pt,twocolumn,letterpaper]{article}

\usepackage{btas}
\usepackage{times}
\usepackage{epsfig}
\usepackage{graphicx}
\usepackage{amsmath}
\usepackage{amssymb}
\usepackage{csquotes}
\usepackage{caption}
\usepackage{url}

\btasfinalcopy 

\ifbtasfinal\fi
\begin{document}

\title{Pushing the Limits of Unconstrained Face Detection:\\ a Challenge Dataset and Baseline Results}

\author{Hajime Nada$^1$,  Vishwanath A. Sindagi$^2$, He Zhang$^3$, Vishal M. Patel$^2$\\
	$^1$Fujitsu Laboratories Ltd., Kanagawa, Japan\\
	$^2$Johns Hopkins University, 3400 N. Charles St, Baltimore, MD 21218, USA\\
	$^3$Rutgers University, 94 Brett Rd, Piscataway Township, NJ 08854, USA\\
	{\tt\small nada.hajime@jp.fujitsu.com, vishwanath.sindagi@gmail.com, he.zhang92@rutgers.edu,}\\ {\tt\small vpatel36@jhu.edu}
}
\maketitle
\thispagestyle{empty}

\begin{abstract}
Face detection has witnessed immense progress in the last few years, with new milestones being surpassed every year. While many challenges such as large variations in scale, pose, appearance are successfully addressed, there still exist several issues which are not specifically captured by existing methods or datasets. In this work, we identify the next set of challenges that requires attention from the research community and collect a new dataset  of face images that involve these issues such as weather-based degradations, motion blur, focus blur and several others. We demonstrate that there is a considerable gap in the performance of state-of-the-art detectors and real-world requirements. Hence, in an attempt to fuel further research in unconstrained face detection, we present a new annotated Unconstrained Face Detection Dataset (UFDD) with several challenges and benchmark recent methods. Additionally, we provide an in-depth analysis of the results and failure cases of these methods. The UFDD dataset as well as baseline results, evaluation code and image source are available at: \url{www.ufdd.info/}

\end{abstract}

\let\svthefootnote\thefootnote
\let\thefootnote\relax\footnote{978-1-5386-7180-1/18 \$31.00 \textsuperscript{\textcopyright}2018 IEEE}
\addtocounter{footnote}{-1}\let\thefootnote\svthefootnote

\section{Introduction}
Face detection is the most important pre-processing step for many facial analysis tasks such as landmark detection\cite{koestinger2011annotated,zhu2012face}, face alignment \cite{zhang2016joint,xiong2013supervised,ren2014face}, face recognition \cite{ranjan2017hyperface}, face synthesis \cite{wang2018high,di2017gp}, etc. The accuracy of face detection systems has a direct impact on these tasks and hence, the success of face detection is of crucial importance. Various challenges such as variations in pose, scale, illumination changes, variety of facial expressions,  occlusion,  \etc, have to be addressed while building face detection algorithms. The success of Viola Jones face detector \cite{viola2001rapid} enabled widespread usage of face detection in a variety of consumer devices and security systems. 

Current state-of-the-art face detectors achieve impressive detection rates on a variety of datasets that contain many challenges. The success of these systems can be attributed to two key steps: (i) advancements in the field of deep learning which has had a direct impact on many facial analysis tasks including face detection, and (ii) dataset collection efforts led by different researchers in the community. Moreover, improvements in detection algorithms have almost always been followed by publication of more challenging datasets and vice versa. Such synchronous advancements in both steps have led to a even more rapid progress in the field. 

\begin{figure*}[ht!]
	\begin{center}		
		
		\includegraphics[width=0.12\linewidth]{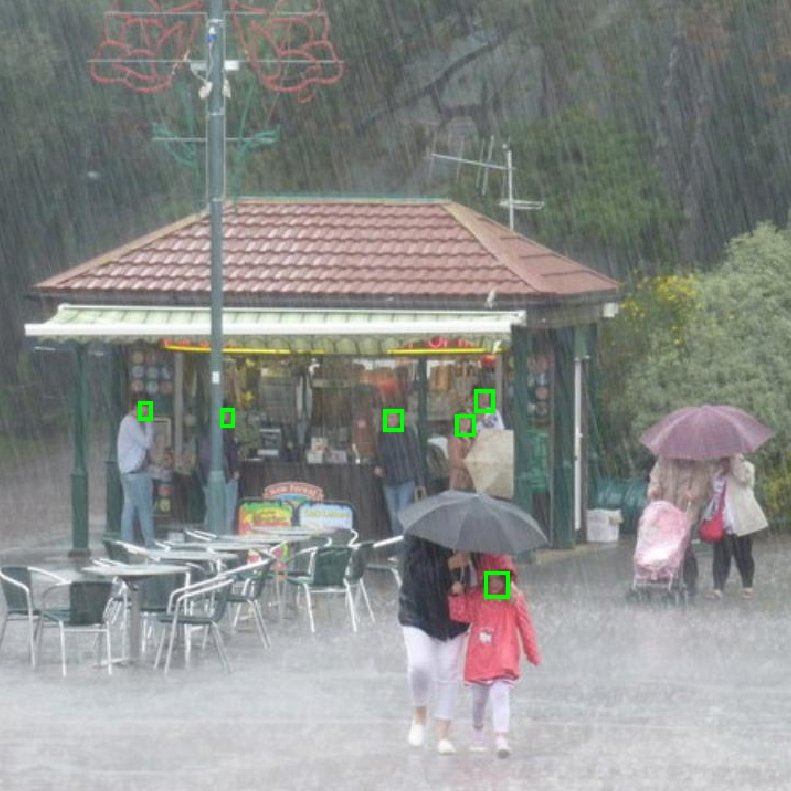}
		\includegraphics[width=0.12\linewidth]{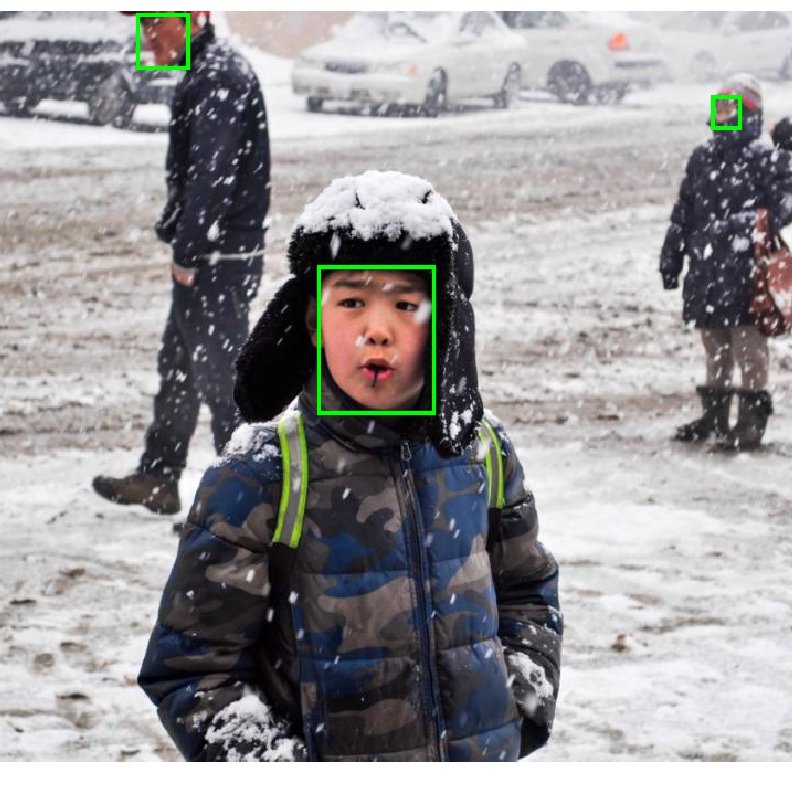}
		\includegraphics[width=0.12\linewidth]{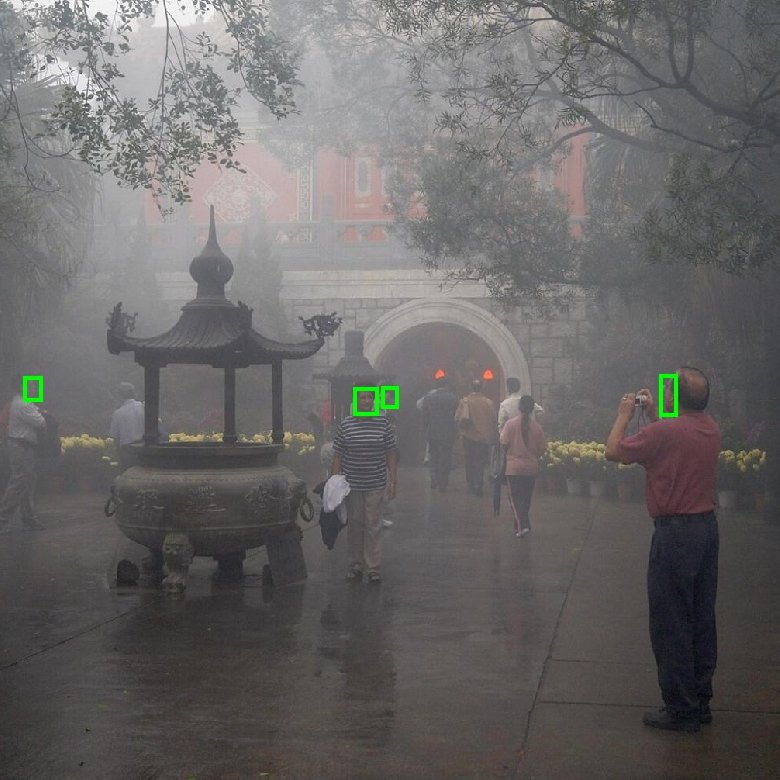}
		\includegraphics[width=0.12\linewidth]{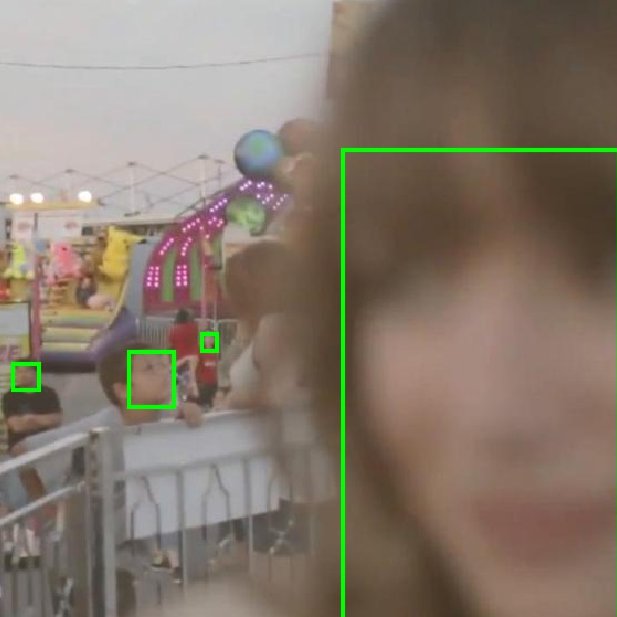}
		\includegraphics[width=0.12\linewidth]{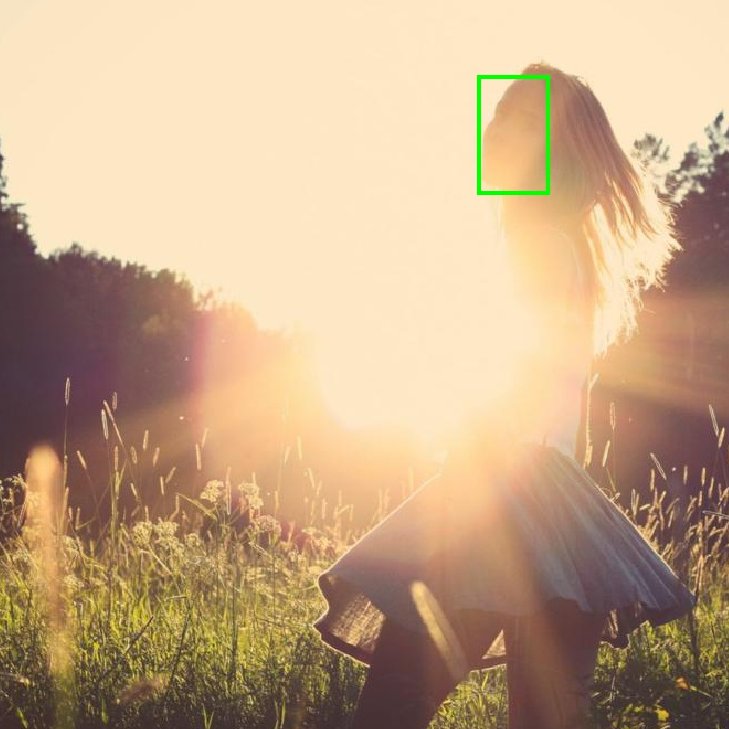}
		\includegraphics[width=0.12\linewidth]{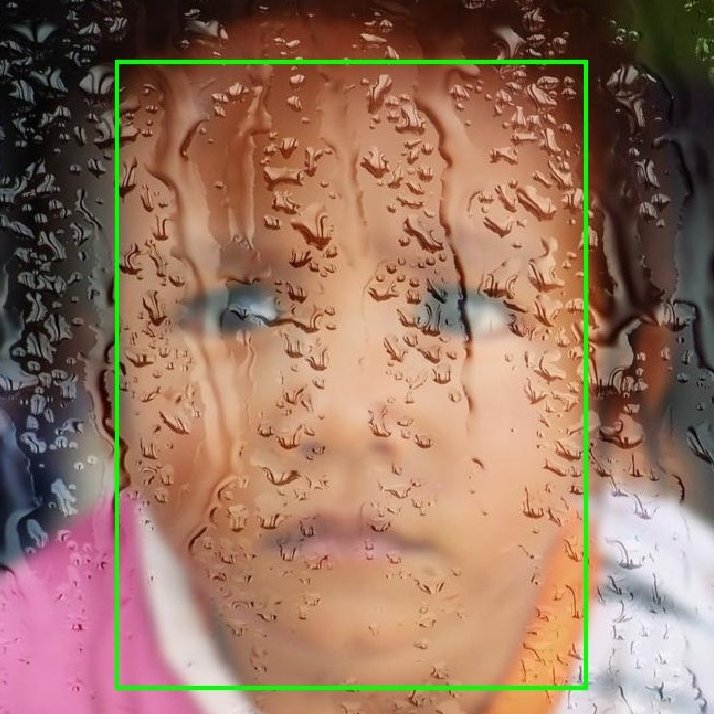}
		\includegraphics[width=0.12\linewidth]{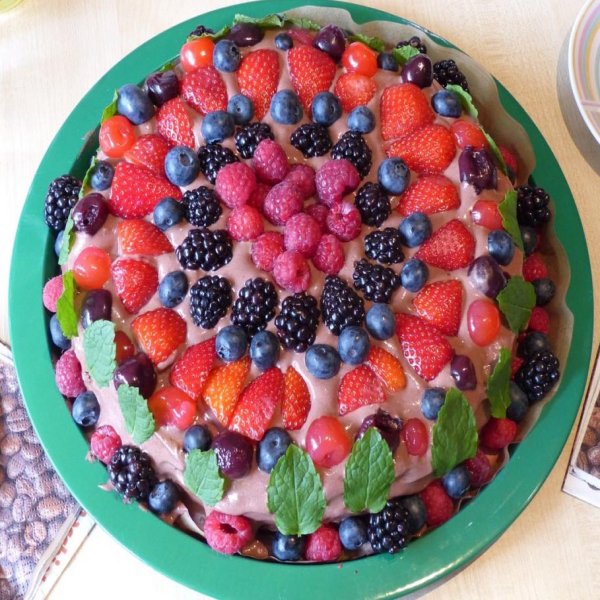}
			
		\includegraphics[width=0.12\linewidth]{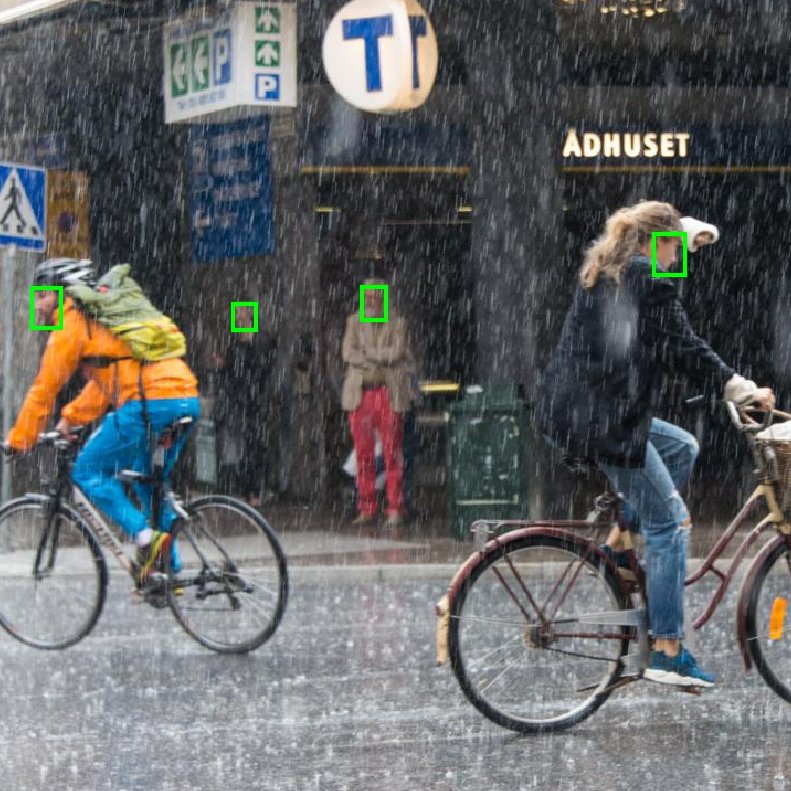}
		\includegraphics[width=0.12\linewidth]{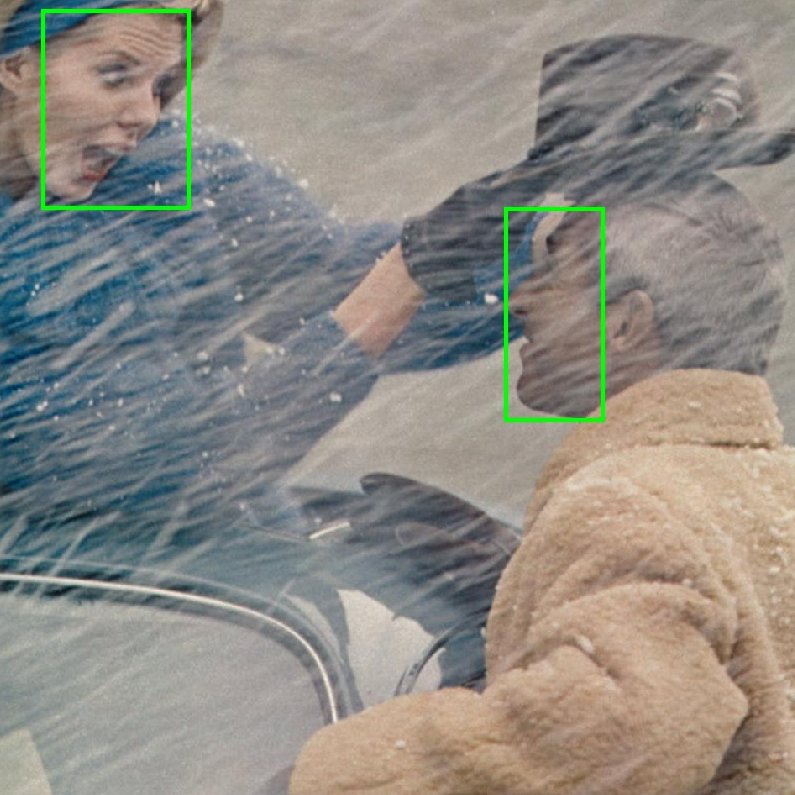}
		\includegraphics[width=0.12\linewidth]{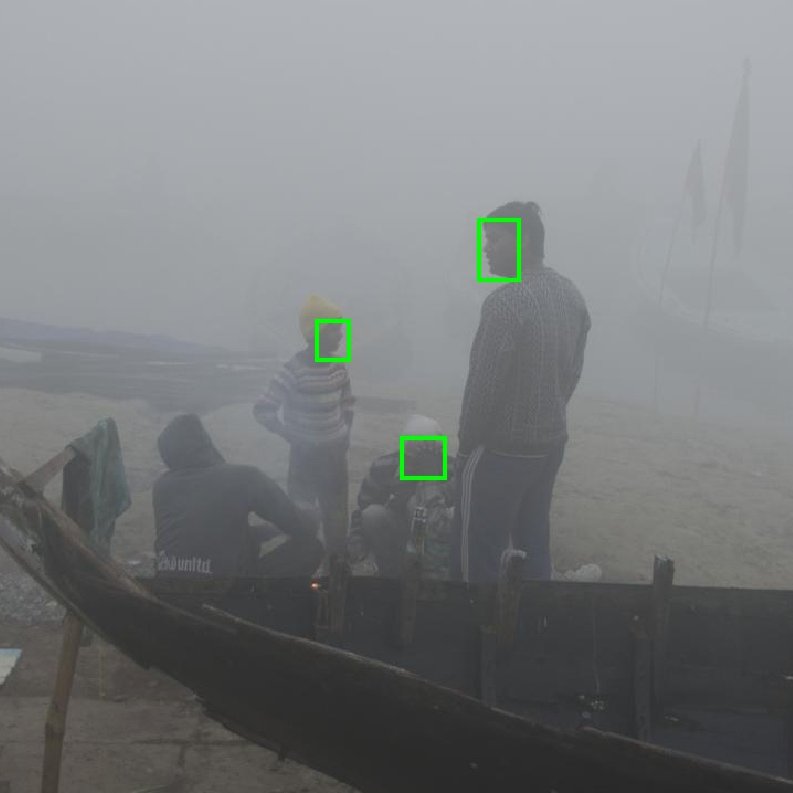}
		\includegraphics[width=0.12\linewidth]{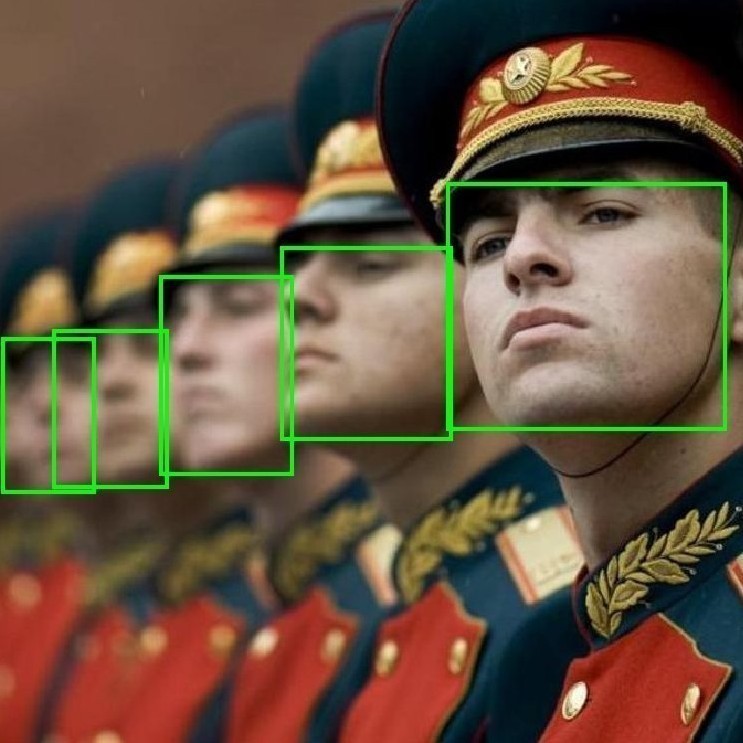}
		\includegraphics[width=0.12\linewidth]{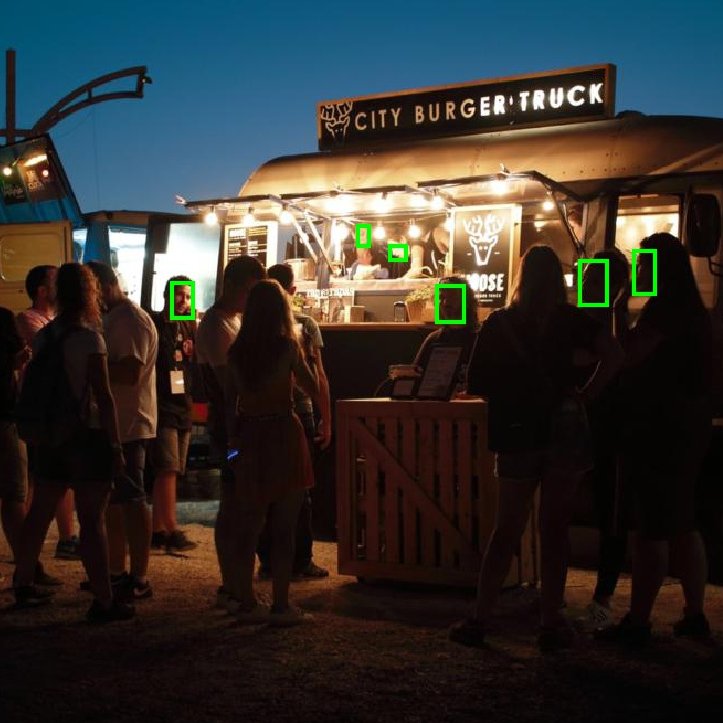}
		\includegraphics[width=0.12\linewidth]{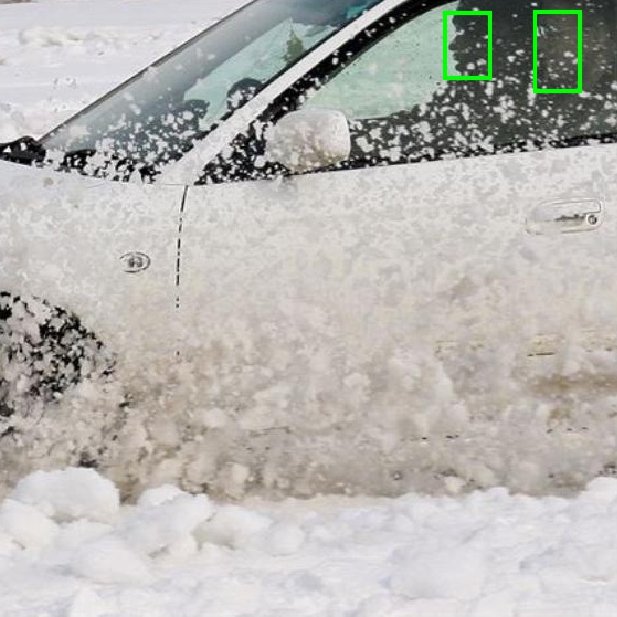}	
		\includegraphics[width=0.12\linewidth]{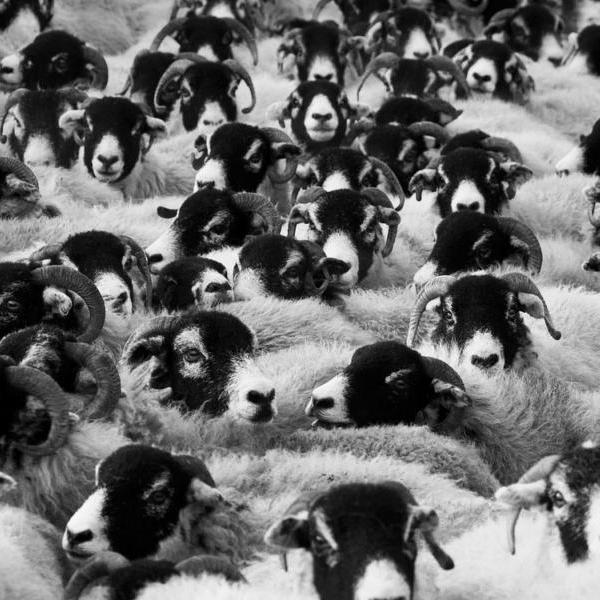}

		\includegraphics[width=0.12\linewidth]{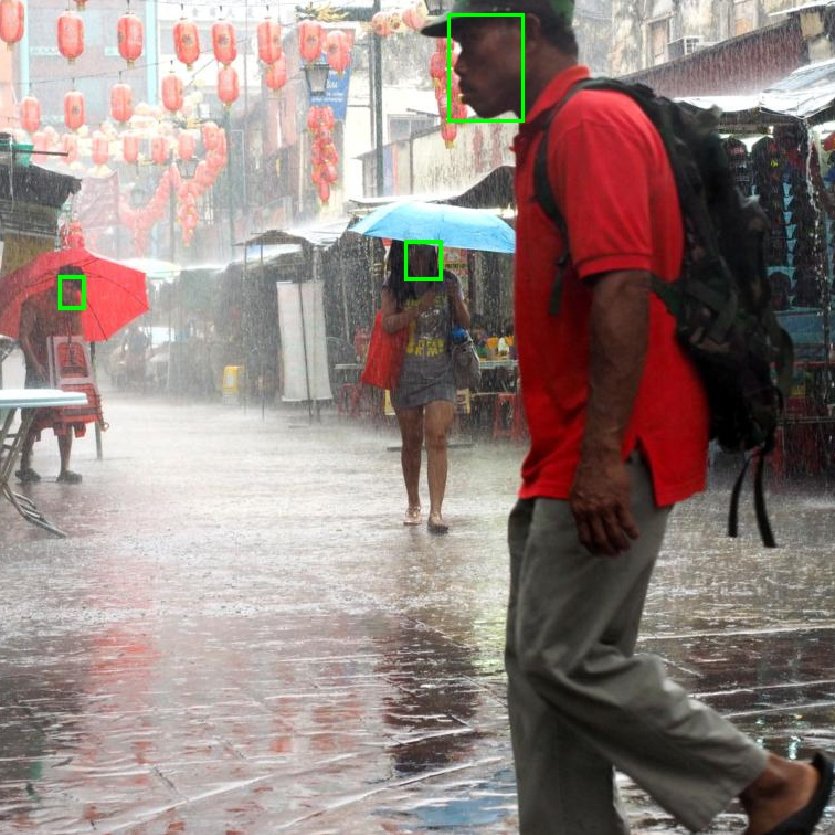}
		\includegraphics[width=0.12\linewidth]{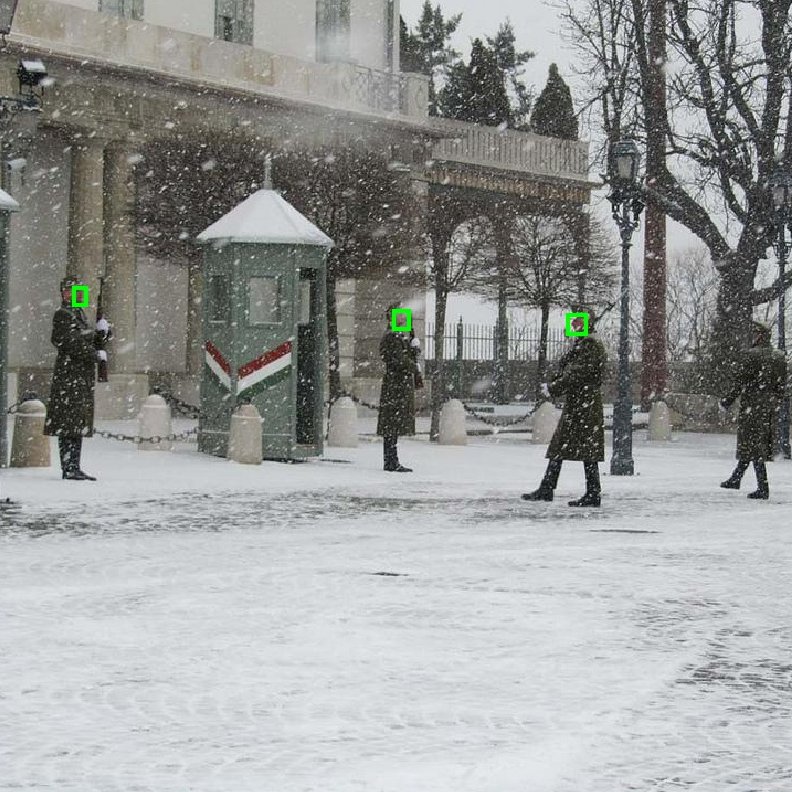}
		\includegraphics[width=0.12\linewidth]{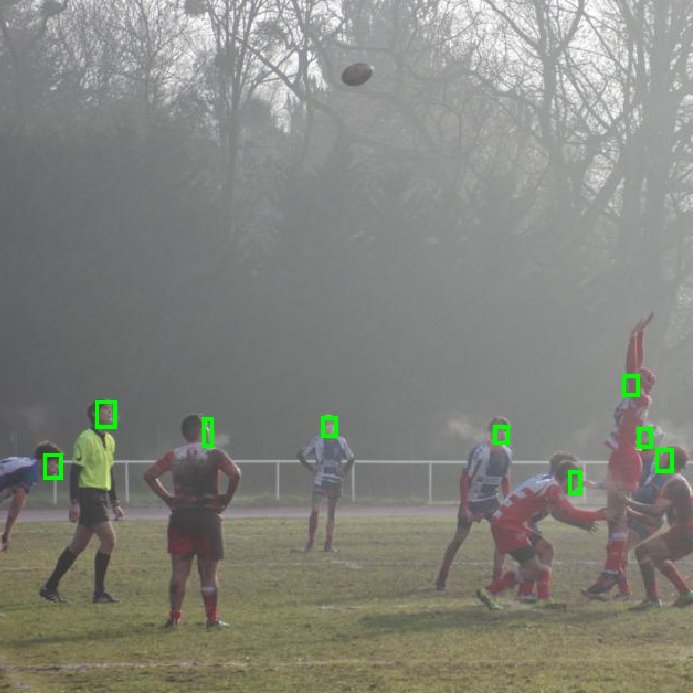}
		\includegraphics[width=0.12\linewidth]{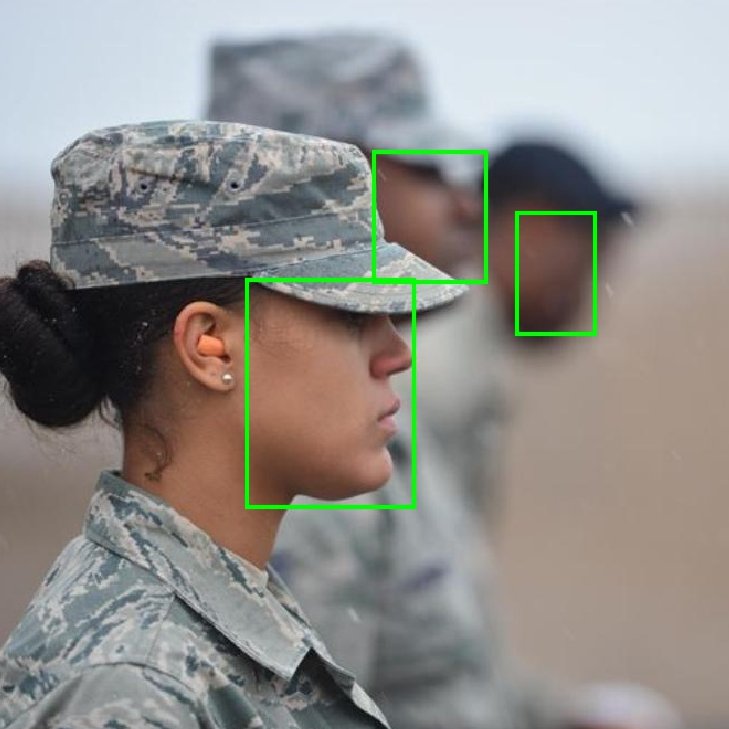}		
		\includegraphics[width=0.12\linewidth]{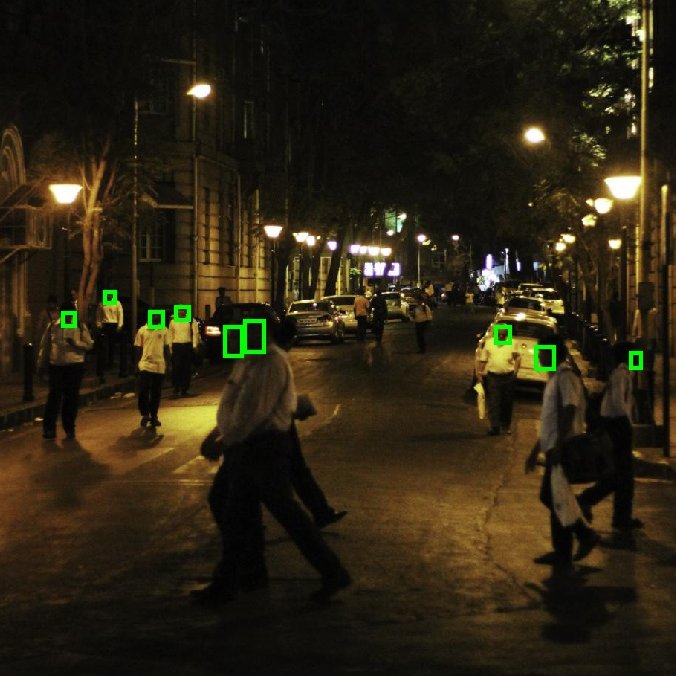}
		\includegraphics[width=0.12\linewidth]{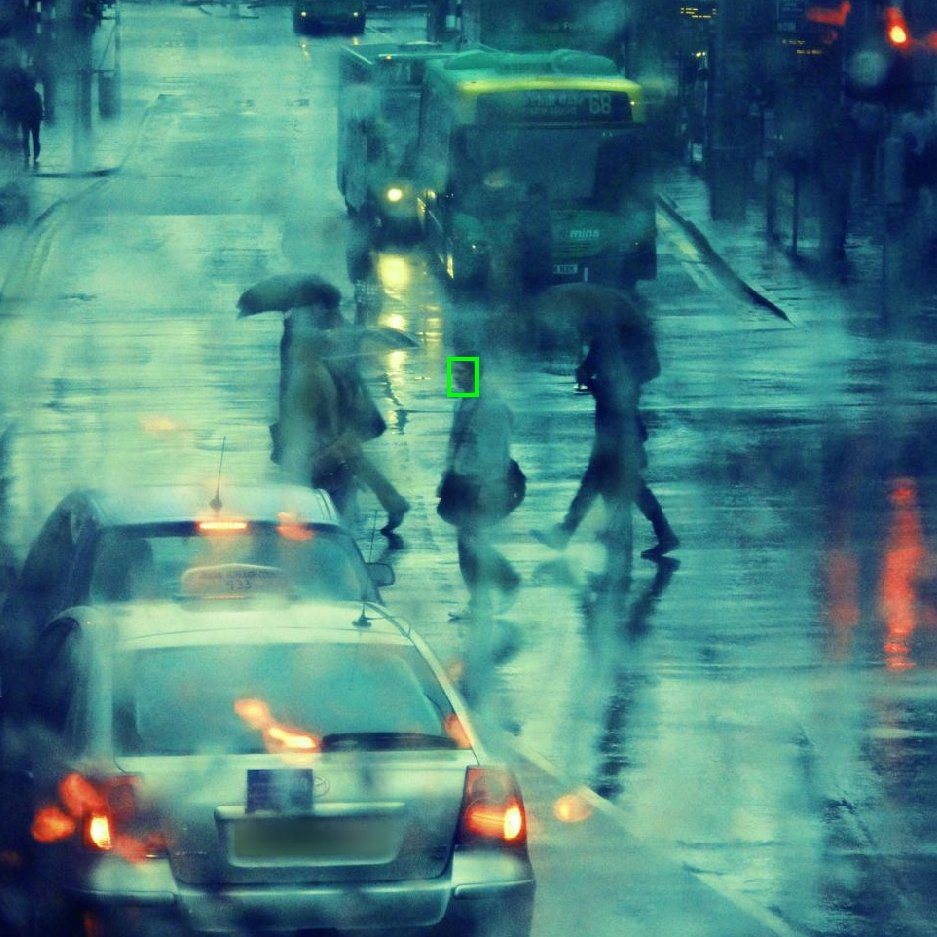}
		\includegraphics[width=0.12\linewidth]{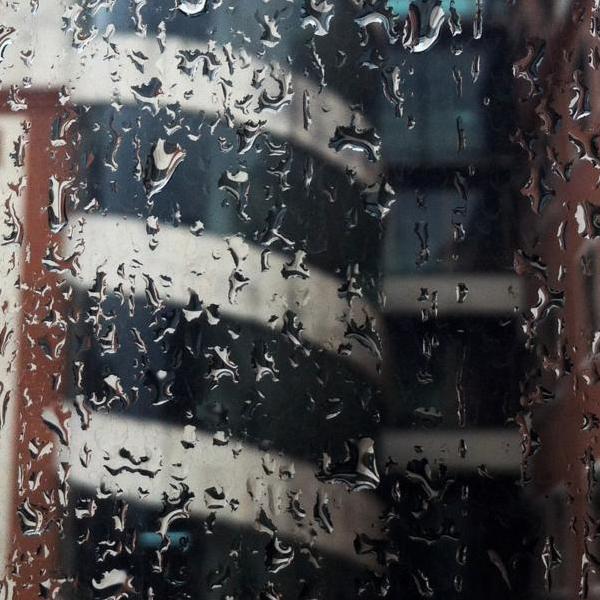}\\

		\hskip 15pt Rain \hskip 40pt Snow \hskip 40pt Haze \hskip 40pt Blur \hskip 30pt Illumination \hskip 5pt Lens impediments \hskip 10pt Distractors
	\end{center}
	\vskip -15pt \caption{Sample annotated images from the proposed UFDD dataset. The dataset is constructed specifically to capture seven different conditions.}
	\label{fig:samples}
\end{figure*}

Various methods have been proposed in the literature that attempt to address different aspects of the face detection problem. Some of the initial work involved design of robust hand-crafted representations \cite{viola2004robust,sung1998example,brubaker2008design,li2014efficient,mathias2014face,yang2014aggregate,chen2014joint} and the use of powerful machine learning algorithms. This was followed by methods \cite{zhu2012face,yan2014face} that exploited structural dependencies in the face. Benchmark datasets such as FDDB \cite{jain2010fddb}, AFW \cite{zhu2012face}, Multi-PIE \cite{gross2010multi} and PASCAL FACE~\cite{everingham2015pascal} enabled the initial progress in face detection research. However, these datasets capture limited variations in scale, pose, occlusion and illumination.

 
In order to address the aforementioned issues, Yang \etal \cite{yang2016wider} presented a large scale face dataset with rich annotations, called as WIDER FACE. They explicitly attempt to capture different variations and demonstrate, through detailed experiments, that a significant gap exists between the accuracy of existing detectors and the expected performance. The performance gap is especially large in the case of tiny faces. Researchers in the community, quickly adopted this new dataset and incorporated advances in CNN-based learning for improving the detection performance. Some of initial works involved cascaded and multi-task CNN-based networks \cite{zhang2016joint}, followed by the use of object detection approaches~\cite{zhu2017cms,jiang2017face} for face detection. Most recent research on face detection has focused more  on improving the performance of small face detection\cite{hu2017finding,zhang2017s3fd,najibi2017ssh}. Some of these approaches employ feature maps from different conv layers \cite{hu2017finding,yang2017face} to build multiple detectors that are robust to scale variations, while others \cite{najibi2017ssh,zhang2017s3fd} have attempted to develop new anchor design strategies to overcome some of the issues faced by anchor design-based methods such as Faster-RCNN \cite{ren2015faster}.

Although WIDER FACE \cite{yang2016wider} attempts to capture a variety of conditions during the dataset collection process, there still exist several practical considerations such as weather-based degradations, different types of blur and distractor images, which have not been explicitly captured by existing face datasets. These conditions are particularly important for a variety of applications such as biometric surveillance,  maritime surveillance, and long-range surveillance, where detection accuracy is of critical importance. Based on this observation, we explore the next set of challenges that require focused attentions from the face detection community, and in this attempt, we present a new Unconstrained Face  Detection Dataset (UFDD) involving a richer set of challenges.

Specifically, this new dataset contains a total of 6,425 images with 10,897 face-annotations and it involves following key degradations or conditions: (1) \emph{Rain}, (2) \emph{Snow}, (3) \emph{Haze}, (4) \emph{Lens impediments}\footnote{By lens impediments, we mean obscurants that appear between the object and camera lens. A few examples are water droplets on the lens, lens dirt, window panes, \etc. }, (5) \emph{Blur}, (6) \emph{Illumination variations}, and (7) \emph{Distractors}. Fig. \ref{fig:samples} shows sample images from the dataset with different variations and its corresponding annotations. For benchmarking existing face detectors, we define two protocols: (1) \textbf{Internal} and (2) \textbf{External}. In the \enquote{internal} protocol, the dataset is divided into $10$-splits and a $10$-fold cross-validation is performed. In the \enquote{external} protocol, the face detectors are trained on another face dataset or a synthetically created face dataset and tested on the real-world dataset. For creating the synthetic dataset, the above listed degradations and conditions are artificially simulated on images from the WIDER FACE dataset. Details of the dataset collection/annotation process and synthetic dataset creation are explained in Section \ref{sec:dataset}.

Through various experiments, we demonstrate that existing algorithms are far from optimal in terms of their detection performance. A  detailed analysis is performed by studying the performance of recent face detectors on this newly proposed dataset. The analysis includes separate study of the effect of different conditions on the performance. In particular, we benchmark four representative algorithms such as Faster R-CNN \cite{ren2015faster}, SSH \cite{najibi2017ssh}, HR-ER \cite{hu2017finding} and S3FD \cite{zhang2017s3fd} using the external protocol and analyze the failure cases of these algorithms. We hope that this detailed analysis will provide deep insights for the design of new algorithms to address these newly identified challenges.

\begin{table*}[ht!]
	\centering
	\caption{Comparison of different datasets. ('\checkmark': Contained in DB. '-': Not contained or mentioned in the paper.)  }
	\label{tab:summary}
\vskip -10pt
	\resizebox{1\linewidth}{!}{%
		\begin{tabular}{|l|c|c|c|c|c|c|c|c|c|c|}
			\hline
			& \#Images & \#Annotations & \multicolumn{8}{c|}{Properties}                                                                           \\ \hline
			&         &                              & Source     & Rain       & Snow       & Haze       & Illumination  & Blur & Lens impediments & Distractors \\ \hline
			AFW \cite{zhu2012face}            & 205     & 473          & Flickr     & -          & -          & -          & -            & -        & -      & -                  \\ \hline
			PASCAL FACE \cite{everingham2015pascal}    & 851     & 1341         & PASCAL-VOC & -          & -          & -          & -            & -          & -           & -           \\ \hline
			FDDB \cite{jain2010fddb}           & 2,845   & 5,171        & Yahoo      & -          & -          & -          & -        & \checkmark   & -               & -           \\ \hline
			MALF \cite{faceevaluation15}            & 5,250   & 11,900       & Flickr, www       & -       & -       & -       & -         & -  & -      & -        \\ \hline
			IJB-C \cite{mazeiarpa-IJB-C}       &  130K  & 300K       & www       & -       & -       & -       & \checkmark    & -      & -        & \checkmark   \\ \hline
			WIDER FACE \cite{yang2016wider}     & 32,303  & 393,703      & www        & \checkmark & -          & -          & \checkmark   & \checkmark     & -     & -           \\ \hline
			UCCS \cite{gunther2017unconstrained}            & -       & 75,738       & camera     & -          & \checkmark & -          & \checkmark   & \checkmark & - & -           \\ \hline
			UFDD (Proposed) & 6,424    & 10,895       & www        & \checkmark & \checkmark & \checkmark & \checkmark   & \checkmark  & \checkmark  & \checkmark  \\ \hline
		\end{tabular}
	}
\end{table*}

\section{Related Work}
\label{sec:relatedwork}

\noindent\textbf{Face detection approaches. }Initial research \cite{viola2004robust,sung1998example,brubaker2008design,li2014efficient,mathias2014face,yang2014aggregate,chen2014joint} on face detection was based on robust hand-crafted representations and involved training of powerful machine learning classifiers. Later approaches such as \cite{zhu2012face,yan2014face} utilized structural dependencies present in faces and modeled them using elastic deformation structures. 

Recently, the success of CNN-based methods in different computer vision tasks such as object recognition \cite{simonyan2014very,he2016deep} and detection \cite{redmon2017yolo9000,ren2015faster,liu2016ssd} has inspired several face detection approaches. Early work on CNN-based face detection involved cascaded architectures \cite{zhang2016joint,li2015convolutional,qin2016joint,yang2015facial} and multi-task training of correlated tasks \cite{ranjan2015deep,ranjan2017hyperface,sindagi2017cnn}. 

Although these approaches were able to obtain impressive detection rates on datasets like Pascal-Faces \cite{yan2014face} and FDDB \cite{jain2010fddb}, the introduction of the WIDER dataset \cite{yang2016wider} demonstrated the lack of robustness of these methods to different factors such as large variations in scales, pose and occlusion. Significant performance gap was observed especially in the case of smaller faces. Hence, most of the recent work involves design of novel strategies to build detectors that are especially robust to scale variations. 
Some methods employ feature maps from multiple layers similar to \cite{zhu2017cms,hu2017finding}, while other methods develop new anchor design strategies \cite{zhang2017s,najibi2017ssh}. Zhu \etal \cite{zhu2017cms} employed the Faster-RCNN  framework and fused features from multiple conv layers to build robustness in scale variations. Additionally, they encoded context to provide additional information to the classifier. Hu \etal \cite{hu2017finding} trained multiple detectors to cater to different scale and employed image pyramids for performing the inference.
Najibi \etal \cite{najibi2017ssh} and  Zhang \etal \cite{zhang2017s} proposed single shot detectors that provided significant improvements while maintaining good computational efficiency. While, Najibi \etal \cite{najibi2017ssh} is based on the region proposal network of Faster-RCNN \cite{ren2015faster}, Zhang \etal \cite{zhang2017s} is based on the SSD \cite{liu2016ssd} detector, where they used additional conv layers converted from VGG-16's fully connected layers. To address the overlap issue of anchor-based techniques, they also introduced new anchor design strategies to ensure increased overlap between anchor boxes and ground-truth faces of smaller sizes during training process. \\

\noindent\textbf{Face detection datasets. } As discussed, face detection is an extensively studied problem. Several datasets such as AFW \cite{zhu2012face}, FDDB \cite{jain2010fddb}, PASCAL FACE \cite{everingham2015pascal}, \etc, have been constructed specifically for face detection.  The AFW  dataset \cite{zhu2012face} consists of 205 images collected from Flickr and has 473 face annotations. Additionally, the authors provide facial landmark and pose labels for each face. The PASCAL FACE dataset \cite{everingham2015pascal} has a total of 851 images which are a subset of the PASCAL VOC  and has a total of 1,341 annotations. These datasets contain only a few hundreds of images and have limited variations in face appearance. 

Jain and Miller in \cite{jain2010fddb} collected a relatively larger dataset that consists of 2,845 images with 5,171 annotations. Although the authors explicitly attempt to capture a wide range of difficulties including occlusions, the images are collected from Yahoo! website and mostly belong to celebrities, due to which the dataset has some inherent bias. The AFLW dataset \cite{koestinger2011annotated} presented a large-scale collection of face images collected from the web, consisting of large variations in appearance, pose, expression, ethnicity, age, gender, \etc. The dataset consists of a total of 25,000 face annotations. However, the AFLW dataset does not have occlusion and pose labels. The IJB-A dataset \cite{klare2015pushing} is constructed for face detection and recognition. It contains 24,327 images with 49,759 face annotations. The recently introduced IJB-C dataset  \cite{mazeiarpa-IJB-C} is an extension of IJB-A with about 138,000 face images, 11,000 face videos, and 10,000 non-face images. The MALF dataset \cite{faceevaluation15} is a large dataset with 5,250 images annotated with multiple facial attributes and it is specifically constructed for fine grained evaluation. 

More recently, Yang \etal \cite{yang2016wider} presented a very large scale dataset called the WIDER FACE with large variations in scale, pose and occlusion. While this dataset demonstrated that existing face detectors performed poorly especially on smaller scale faces, recent CNN-based face detectors \cite{najibi2017ssh,zhang2017s} have incorporated robustness to scale variations and have achieved impressive performances. Additionally, these datasets do not focus on specifically capturing weather-based degradations such as snow, rain, haze, \etc. Gunther \etal \cite{gunther2017unconstrained} recently presented an unconstrained dataset for face detection and recognition, where the authors do attempt to capture weather-based degradations, however, these are limited to a smaller set of conditions such as sunny day and snowy day. Several conditions such as haze and rain are not captured. In contrast, the proposed dataset in this work captures a much larger set of variations with large set of images in each condition. Additionally, we also include a large set of distractor images which is largely ignored by the existing datasets. Table \ref{tab:summary} gives a summary of different datasets in comparison with the proposed dataset.

\section{Dataset}
\label{sec:dataset}


To the best of our knowledge, UFDD is among the first datasets that explicitly captures variations in different weather conditions and other degradations such as lens impediments, motion blur and focus blur. Additionally, the dataset also consists of a large set of distractor images which is largely ignored by the existing datasets where every image almost necessarily has at least one face annotation. These images either contain non-human faces such as animal faces or no faces at all. The presence of distractor images is especially important to measure the performance of a face detector in rejecting non-face images and to study the false positive rate of an algorithm. 

Some existing datasets capture a few of these conditions separately. For instance, the UCCS dataset \cite{gunther2017unconstrained} contains sunny, snow and blur images, however, other degradations such as haze and rain are missing. Moreover, these images are collected from a single location using a surveillance camera. In contrast, the proposed dataset is collected from the Internet and hence, it is more diverse.  As a result, this dataset can be used to evaluate the generalization ability of different face detectors on a diverse set of conditions.

Similar to FDDB \cite{jain2010fddb}, we define two separate protocols to evaluate face detection performance on the proposed dataset:  (1) \textbf{Internal} and (2) \textbf{External}. In the \enquote{internal} protocol, the dataset is divided into $10$-splits and a $10$-fold cross-validation is performed. In the \enquote{external} protocol, the face detectors are trained on another face dataset or on a synthetically created face dataset and tested on the real-world dataset. In this work, we use the WIDER FACE dataset as another training dataset to create the synthetic dataset.

\subsection{Data Collection and Annotation}

\noindent\textbf{Collection and Annotation. } Images in the proposed dataset are collected from different sources on the web such as Google, Bing, Yahoo, Creative commons search, Pixabay, Pixels, Wikimedia commons, Flickr, Unsplash, Vimeo and Baidu. Images are searched using various keywords such as \enquote{rain + faces}, \enquote{snow + faces}, \enquote{rain + crowd}, \enquote{dark + crowd} \etc. Images are collected in such a way that the dataset captures a total of seven conditions and degradations. These conditions are chosen based on the observation that they are entirely plausible in a variety of applications such as video surveillance and maritime surveillance. In section \ref{sec:evaluation}, we present a detailed analysis of the effect of these conditions separately on face detection performance. 

Wherever possible, we ensured a uniform distribution of different conditions so that the dataset has minimal bias towards any particular condition. Table~\ref{tab:distribution} shows the distribution of number of images per condition.

\begin{table}[t!]
		\centering
		\caption{Distribution of images in the UFDD dataset.}
		\label{tab:distribution}
		\vskip -10pt
		\small
	\resizebox{1\linewidth}{!}{%
	\begin{tabular}{|l|c|c|c|c|c|c|c|}
		\hline Condition & Rain & Snow & Haze & Blur & Illumination & \begin{tabular}[c]{@{}l@{}}Lens\\ impediments\end{tabular} & Distractors \\ \hline
		\#Images  & 628  & 680  & 442  & 517  & 612          & 95                                                         & 3450       \\ \hline
	\end{tabular}
}
\end{table}

After collection, the dataset is cleaned to remove near-duplicate images using \cite{hash}. After dataset cleaning, the images are resized to have a width of $1024$ while preserving its original aspect ratio. These resized images are used for annotation and evaluation. For annotations, these images are uploaded to Amazon mechanical turk (AMT) and each image is assigned to around $5$ to $9$  AMT workers. The workers are asked to annotate all recognizable faces in the image. Once the annotation is complete, the labels are then cleaned and consolidated using the procedure outlined by Taborsky \etal in \cite{taborsky2015annotating}. The consolidation process is viewed as a clustering problem. For each image, annotations from all workers are converted to a list of sets, where each set represents annotations of a particular face in the image by different workers. For instance, if there are $n$ faces in an image, then there would be $n$ sets in the final list. First, sets in the list are initialized by annotations created by the first worker. Annotations from other workers are then added to those sets if the overlap between them is greater than $0.3$, otherwise a new set is created. After processing all annotations, the list consist of sets of overlapping annotations, ideally corresponding to each face in the image. Final, a pruning step is carried out to remove erroneous annotations, where sets from the final list are removed if they do not contain at least 2 boxes. For each remaining set in the list, the average bounding box is computed and used as the ground-truth for the image.

\section{Evaluation, Benchmarking and Analysis}
\label{sec:evaluation}

\subsection{Methods used for evaluation}

We evaluate the following recent face detection approaches on the proposed UFDD dataset. 

\noindent\textbf{Faster-RCNN. } Faster-RCNN \cite{ren2015faster} is among the first end-to-end CNN-based object detection methods and it consists of a region proposal network (RPN) and a region classification network (RCN). RPN is based on VGG-16 \cite{simonyan2014very} architecture and produces candidate regions which are agnostic to object class. These candidate regions are further processed by RCN which pools features from the final conv layer of VGG-16 and forwards them through a set of fully connected layers to produce the final object class and bounding box. Since most face detectors are based on anchor boxes and Faster-RCNN was the first method to propose anchor boxes, this method was chosen to be the baseline approach. For the purpose of evaluation, we used an open-source implementation \cite{pyfaster} specifically implemented for face detection which is based on the original Faster-RCNN source code. 


\noindent\textbf{HR-ER. }Hu \etal \cite{hu2017finding} specifically addressed the problem of large variations in scale found in the WIDER FACE dataset by designing scale-specific detectors based on ResNet-101 \cite{he2016deep}. Each scale-specific detector is a conv layer which processes features extracted from earlier layers and produces a spatial heat map that indicates the detection confidence at every location. Image pyramids are used during training and inference. Additionally, balanced-sampling  and  hard negative mining are employed during training to effectively learn difficult samples.  For the purpose of evaluation, we used the implementation provided by the authors. 

\noindent\textbf{SSH. } Najibi \etal \cite{najibi2017ssh} presented a single stage headless (SSH) face detector, which is primarily based on the RPN of Faster-RCNN. In contrast to Faster-RCNN, SSH consists of multiple detectors placed on top of different conv layers of VGG-16 to explicitly address scale variations. Each detector is designed with an additional context processing module that incorporates surrounding context by increasing the receptive filed of the network. In contrast to earlier multi-scale detection work \cite{cai2016unified}, the authors fuse features from different layer before using them for detection. Also, similar to \cite{shrivastava2016training}, they use online hard example mining to boost the detection performance. For the purpose of evaluation, we used the implementation provided by the authors.

\noindent\textbf{S3FD. } Similar to \cite{najibi2017ssh}, Zhang \etal \cite{zhang2017s} proposed single shot scale invariant face detector (S3FD) where they presented new anchor design strategies to overcome issues of anchor-based techniques for small object detection. 
The authors propose a max-out background technique to address the issue of high false positive rate of small faces. S3FD is based on the popular object detection framework called single shot detector (SSD) \cite{liu2016ssd}, where they use VGG-16 as the base network. Similar to earlier approaches on face detection and object detection, S3FD uses hard negative mining to improve the detection accuracy. For the purpose of evaluation, we used the implementation provided by the authors.


\begin{figure}[htp!]
	\begin{center}				
		\includegraphics[width=.55\linewidth, height=0.55\linewidth]{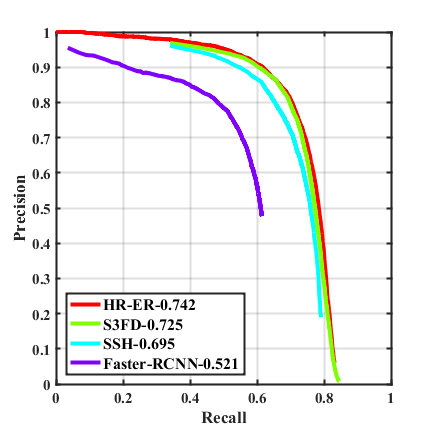}
	\end{center}
	\vskip -15pt \caption{Evaluation results of different algorithms, that are pre-trained on WIDER FACE, on the proposed UFDD dataset.} 
	\label{fig:pretrained_pr_curve}
\end{figure}

\subsection{Evaluation and Analysis}
For the purpose of analysis, the aforementioned methods are evaluated in two different scenarios: 

\noindent(i) Using pre-trained models: As argued by Yang \etal in \cite{yang2016wider}, most of the recent methods (including the ones described above) use WIDER FACE as a source dataset as it is a significantly large dataset that captures large scale variations in different factors and conditions. Based on this argument, we evaluate these recent methods which are pre-trained on WIDER FACE directly on the proposed UFDD dataset. Fig. \ref{fig:pretrained_pr_curve} shows the precision-recall curves corresponding to various methods, that are pre-trained using the WIDER FACE training set \cite{yang2016wider}, on the proposed UFDD dataset.  Contrary to the suggestions made by the authors of \cite{yang2016wider}, it can be observed that WIDER FACE need not necessarily be effective as a source training set especially for constraints such as rain, snow, haze, blur, distractors, that are captured by the UFDD dataset. The poor performance of state-of-the-art detectors in such cases highlights the need for a dataset that explicitly captures them. Additionally, this argument calls for an improvement on the design of algorithms and networks in order to capture these kind of variations and hence, improve the robustness of the detectors. 

\begin{figure}[htp!]
	\begin{center}
		\includegraphics[width=0.25\linewidth]{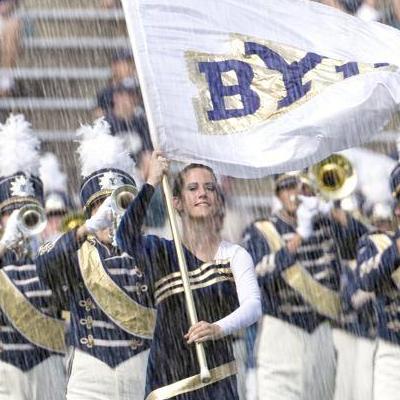}
		\includegraphics[width=0.25\linewidth]{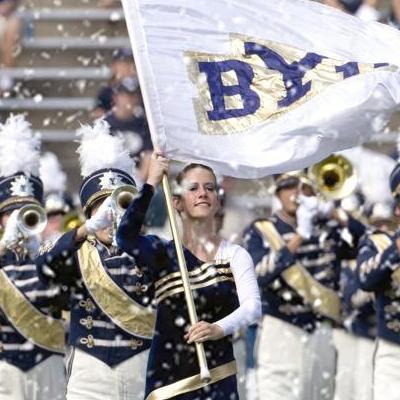}
		\includegraphics[width=0.25\linewidth]{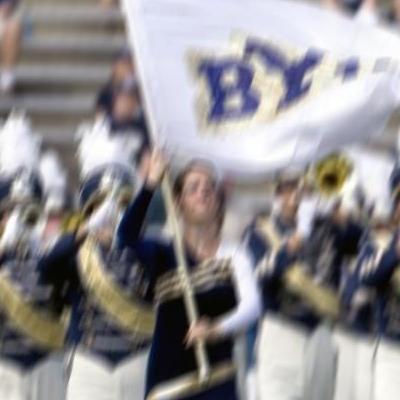}
		\vskip +2pt
		\includegraphics[width=0.25\linewidth]{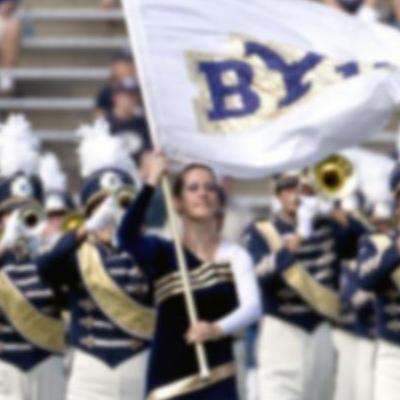}
		\includegraphics[width=0.25\linewidth]{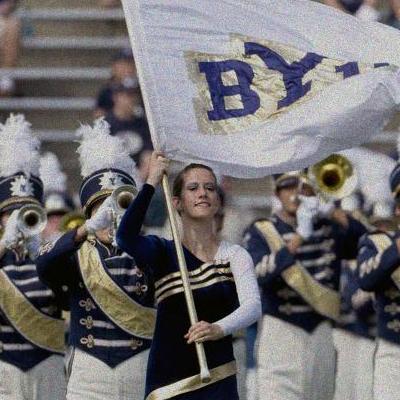}
		\includegraphics[width=0.25\linewidth]{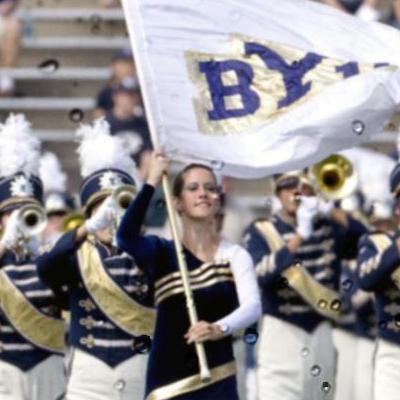}
	\end{center}
	\vskip -15pt \caption{Sample annotated images from the synthetic WIDER FACE dataset. Left to right and top to bottom: Rain, snow, motion blur, Gaussian blur, illumination, lens impediments.}
	\label{fig:samples_synthetic}
\end{figure}

\noindent(ii) Use synthetic dataset for fine-tuning: Under ideal conditions, one would want to train their networks on large scale datasets. However, conditions such as rain, snow, and haze occur with relatively less frequency, due to which the availability of such images on the web is limited. A potential solution to address this issue is to synthesize these conditions and simulate images containing these less frequently occurred constraints. Since WIDER FACE \cite{yang2016wider} is the largest face dataset containing occlusions, different scale variations, and difficult poses, we use this dataset to produce the synthetic dataset that contains variations such as rain, snow, lens impediments and blur \footnote{Transmission maps are required to synthesize hazy images \cite{he2011single,ren2016single}.  Since transmission maps are not available in the considered datasets, we are unable to synthesize the corresponding hazy images.}. Fig. \ref{fig:samples_synthetic} illustrates sample images from the synthetic dataset. In the following, we discuss the details of the synthesis procedure for different conditions.

\noindent \textbf{Rain:} Following \cite{synth-rain},  15 large rainy masks are synthesized, which are used to be  blended with the images in WIDER FACE \cite{yang2016wider} to synthesize rainy images. Particularly,  these 15 masks are  with three Gaussian noise levels ($16$, $32$, and $48$), and five rotation angles ($70^\circ, 80^\circ, 90^\circ, 100^\circ$ and  $110^\circ$). 

\noindent \textbf{Snow:} Following the procedure discussed in \cite{synth-snow}, 3 large snowy masks are synthesized. In particular,  3 different masks with different resize ratios and number of mixtures, $(75\%, 9), (100\%, 12)$ and $(133\%, 16)$ are synthesized. Finally we get 15 types of snowy masks with same rotation method as rainy image. Then we crop mask randomly and blend it with the original image.

\noindent \textbf{Blur:} Both focus blur and motion blur are used to synthesize the blurry images. Particularly,  3 levels of focus blur kernels ($\alpha, 1.5\alpha$ and $2\alpha$, where $\alpha =$ image height $/ 640$) are leveraged to synthesize images with focus blur and 3 levels of motion blur kernels ($5\beta, 10\beta$ and $15\beta$, where $\beta =$ image height $/ 640$) are used to synthesize images with motion blur. We use random motion angles in the range of  $[0^\circ , 180^\circ]$.

\noindent \textbf{Illumination:} Pixel intensity values of the original images are modified to make these images brighter or darker. To make these images brighter, we change the image intensity from $[0, 255\gamma]$ to $[0, 255]$, where $\gamma = 0.6\delta + 0.4$ and $\delta$ is random number in $(0, 1)$. To make these images darker, we add Possion noise  with \cite{matpoisson} to reproduce (or approximate) shot noise by high ISO sensitivity as discussed in \cite{noise-iso} and change the image intensity from $[0, 255]$ to $[0, 255\gamma]$, where $\gamma$ is the same as above.

\noindent \textbf{Lens impediments} Seven different lens impediment masks are downloaded from web and are  blended with the images by the procedure discussed in \cite{synth-lens}. The masks have almost uniform background to keep the contrast of images. To increase the number of masks, we rotate, combine and crop each mask to be similar in size as the image. Then, these augmented masks are blended with images with opacity in the range of  $[0.5, 1]$.

\begin{figure}[htp!]
	\begin{center}				
		\includegraphics[width=.55\linewidth, height=0.55\linewidth]{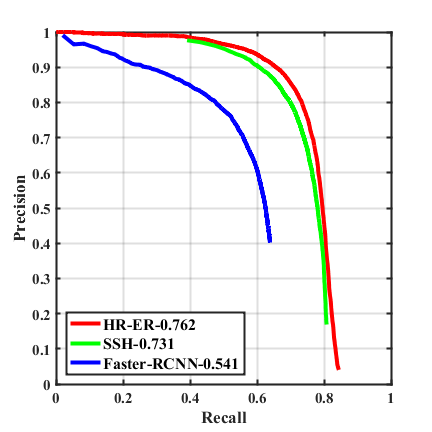}	
	\end{center}
	\vskip -15pt \caption{Evaluation results of different algorithms on the proposed UFDD dataset. Note that the face detectors are trained on the synthetic WIDER FACE dataset.} 
	\label{fig:synthetic_pr_curve}
\end{figure}

This synthetic dataset is then used as a source training dataset to fine-tune the existing state-of-the-art face detectors discussed above. Fig.~\ref{fig:synthetic_pr_curve} shows the precision-recall curves corresponding to different methods (trained on the proposed synthetic dataset) evaluated on the proposed UFDD datasets. Table \ref{tab:map_source} shows the mean average precision (mAP) corresponding to different methods that are trained on the original WIDER FACE training set and synthetic dataset. It can be observed that there is considerable improvements in the detection performance when the networks are trained on the synthesized dataset. This also demonstrates the limitations of existing large scale face detection datasets, where many real-world conditions such as rain and haze are not considered.  

\begin{table}[htp!]
	\centering
	\caption{The mAP scores using different training sets.}
	\label{tab:map_source}
	\vskip -10pt
	\resizebox{1\linewidth}{!}{%
		\begin{tabular}{|l|c|c|}
			\hline
			Training set& Original WIDER FACE & Synthetic WIDER FACE \\ \hline
			Faster-RCNN \cite{ren2015faster} & 0.521               & 0.541                \\ \hline
			SSH   \cite{najibi2017ssh}      & 0.695               & 0.731                \\ \hline
			S3FD   \cite{najibi2017ssh}     & 0.725               & -                  \\ \hline
			HR-ER  \cite{hu2017finding}     & 0.742               & 0.762                \\ \hline
		\end{tabular}
	}
\end{table}

\begin{figure*}[ht!]
	\begin{center}		
		
		\centering\rotatebox{90}{Annotation}
		\includegraphics[width=0.13\linewidth]{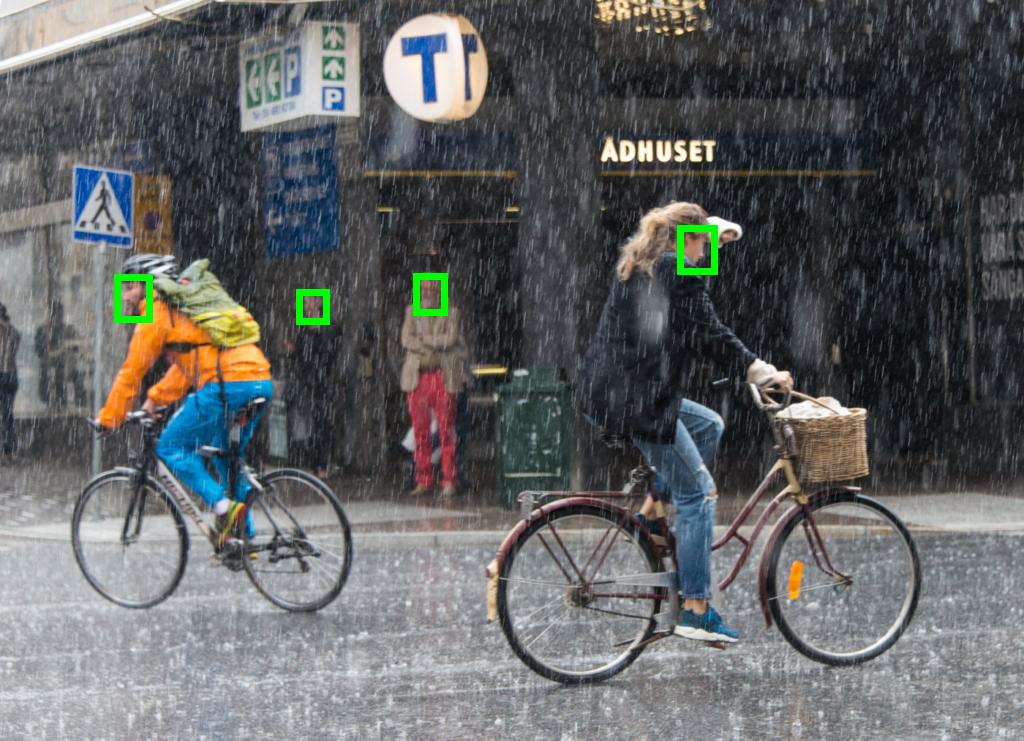}
		\includegraphics[width=0.13\linewidth]{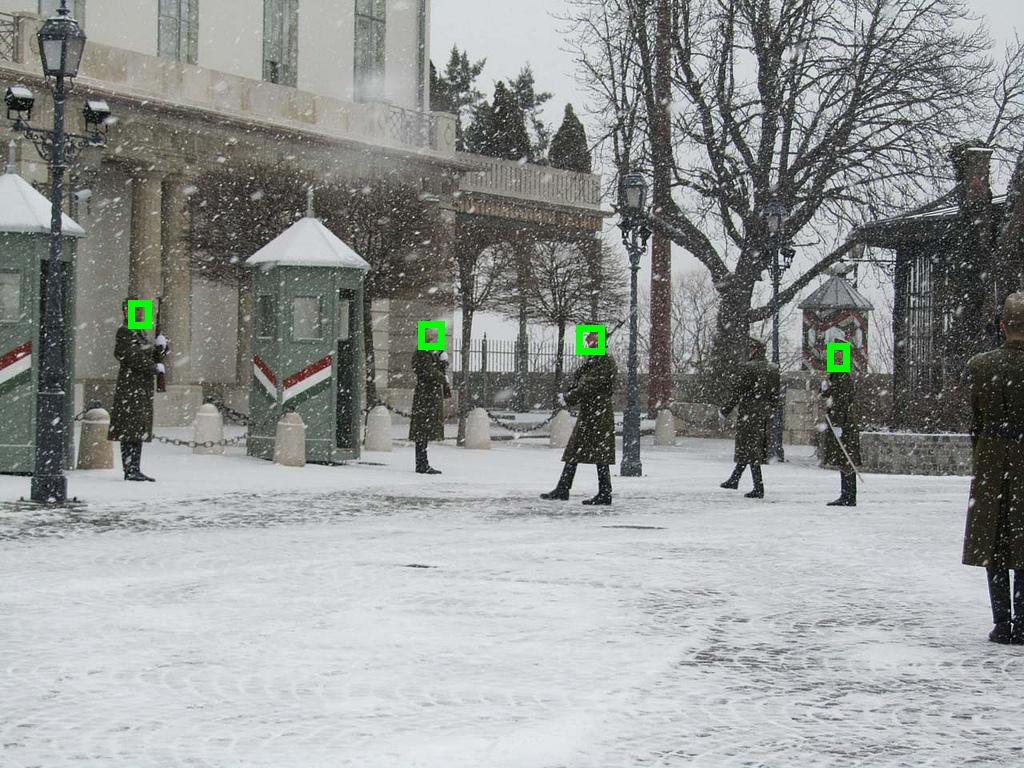}
		\includegraphics[width=0.13\linewidth]{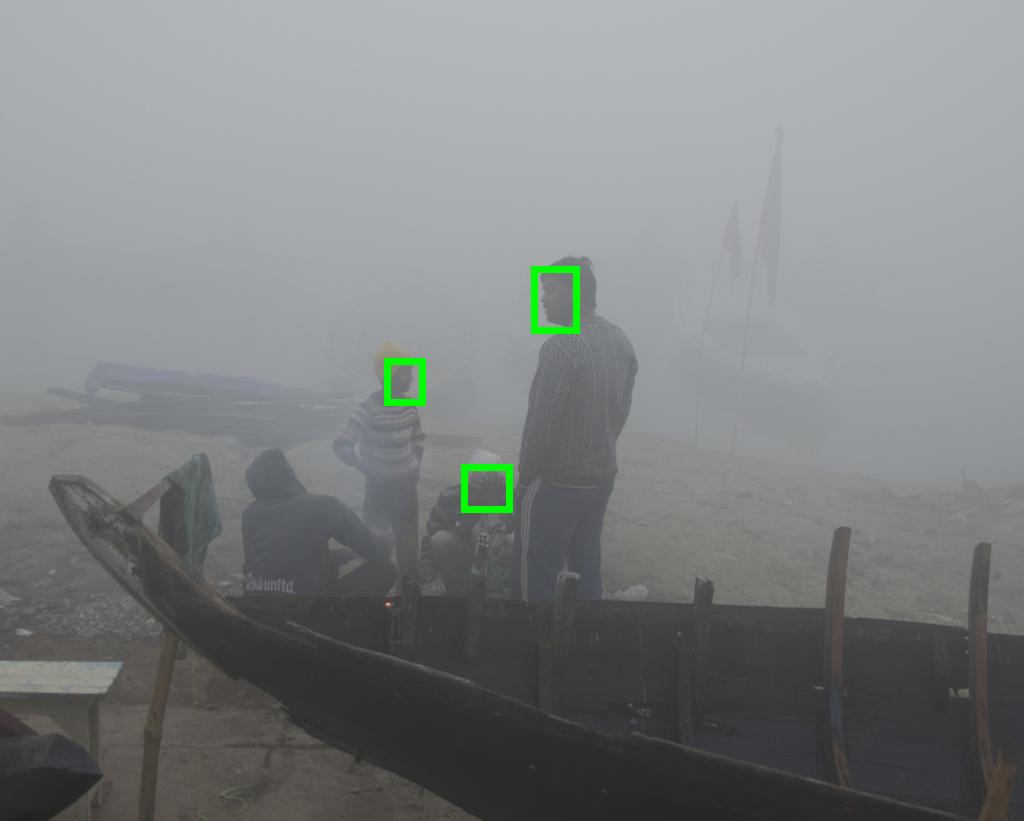}
		\includegraphics[width=0.13\linewidth]{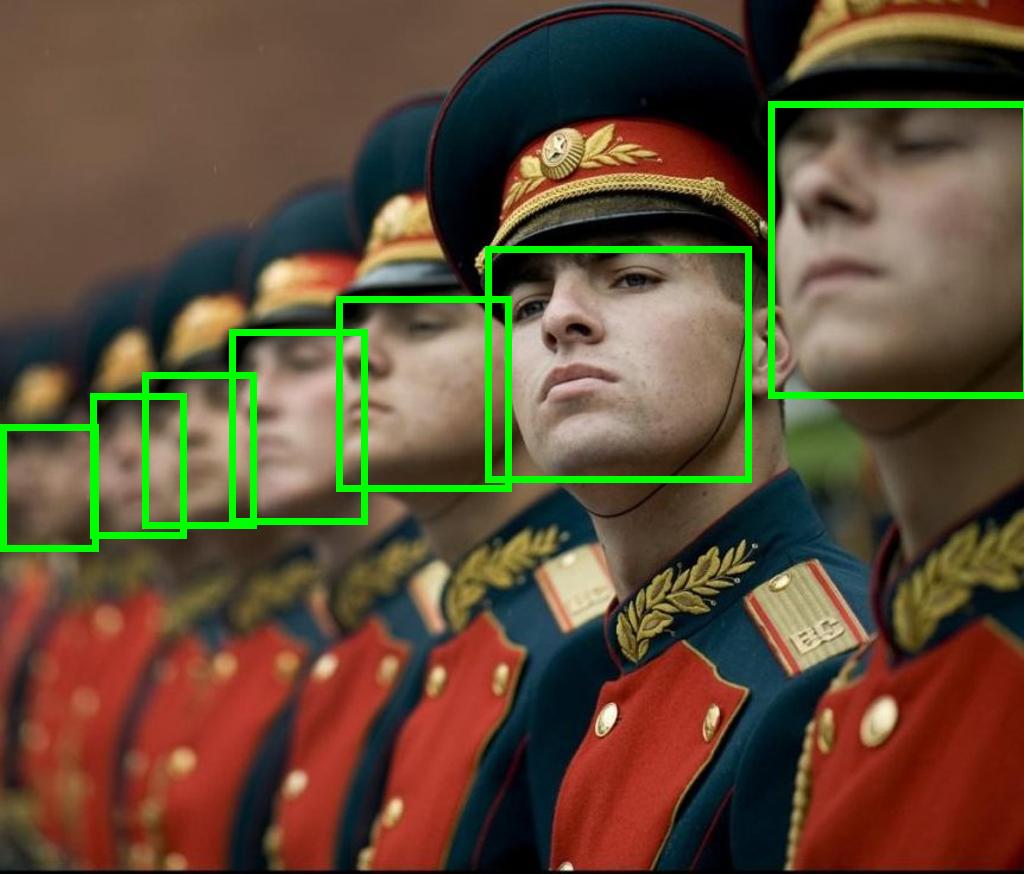}
		\includegraphics[width=0.13\linewidth]{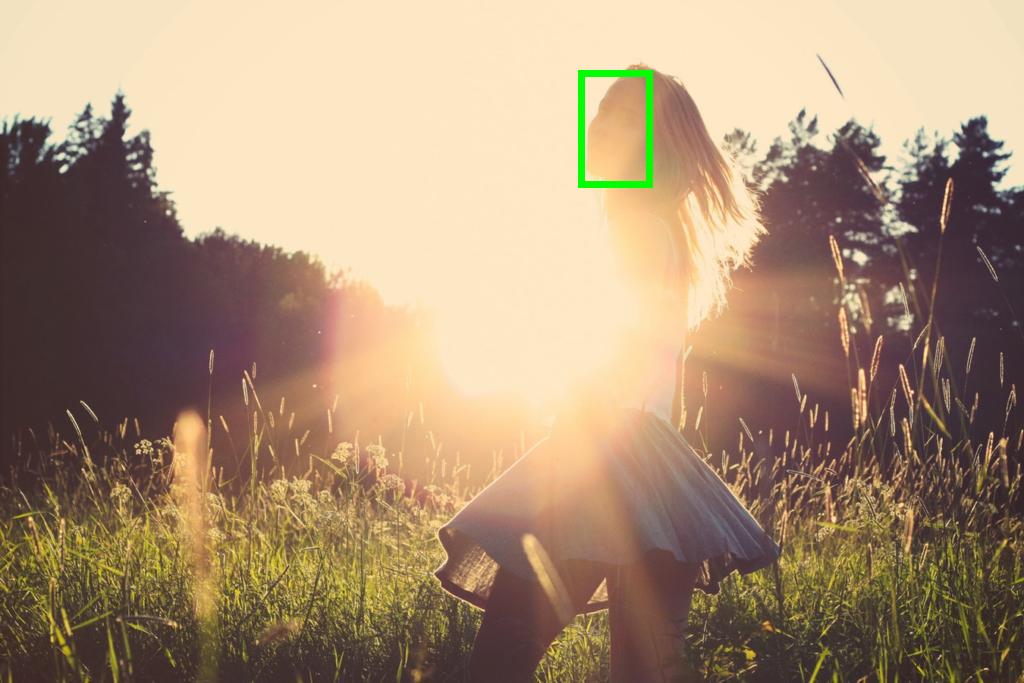}
		\includegraphics[width=0.13\linewidth]{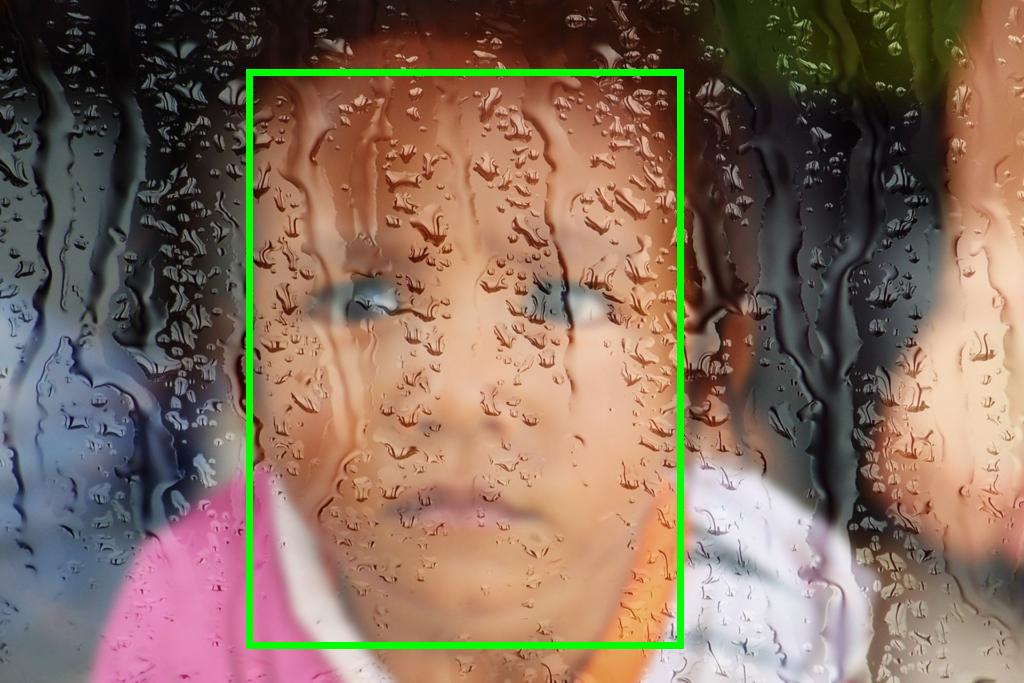}
		\includegraphics[width=0.13\linewidth]{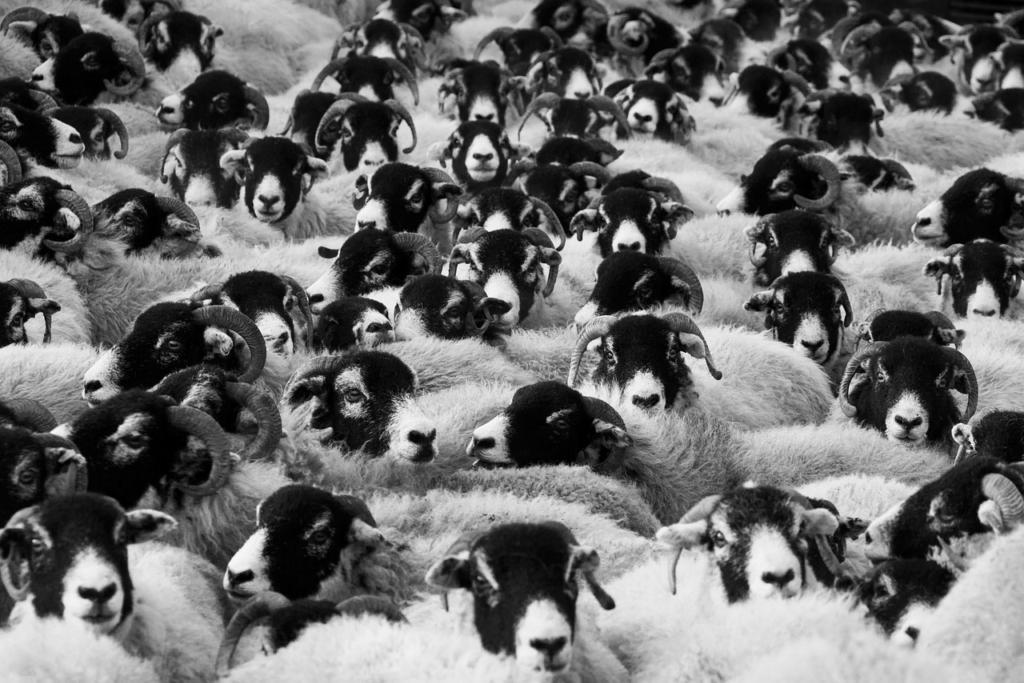}
		
		\rotatebox{90}{Faster-RCNN \hskip 15pt}
		\includegraphics[width=0.13\linewidth]{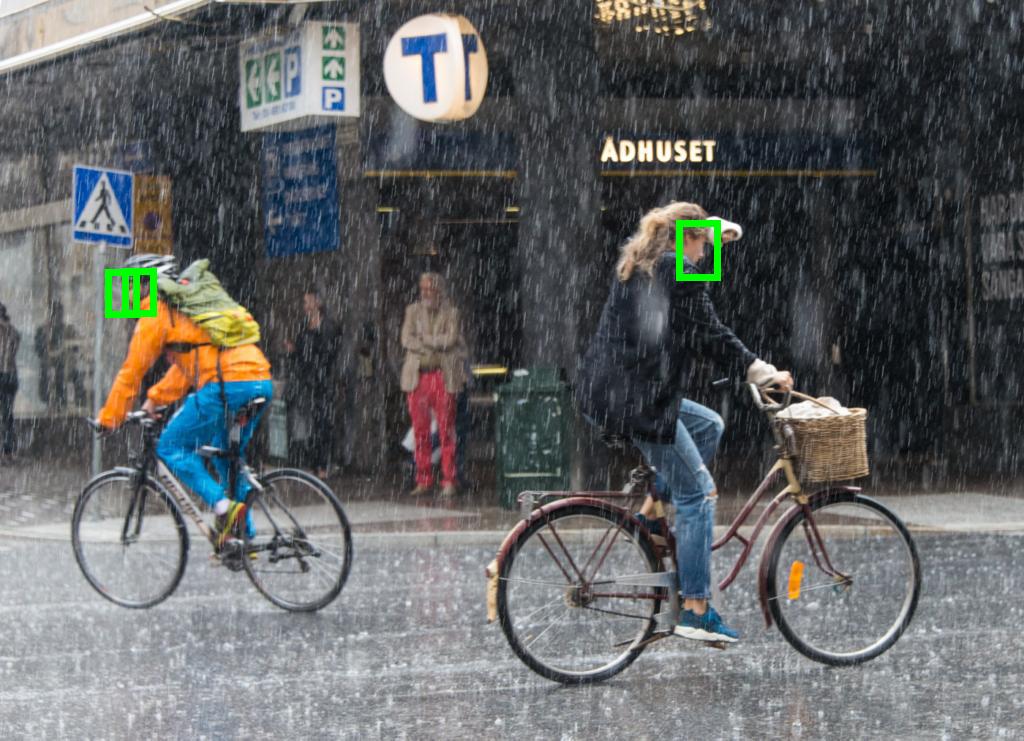}
		\includegraphics[width=0.13\linewidth]{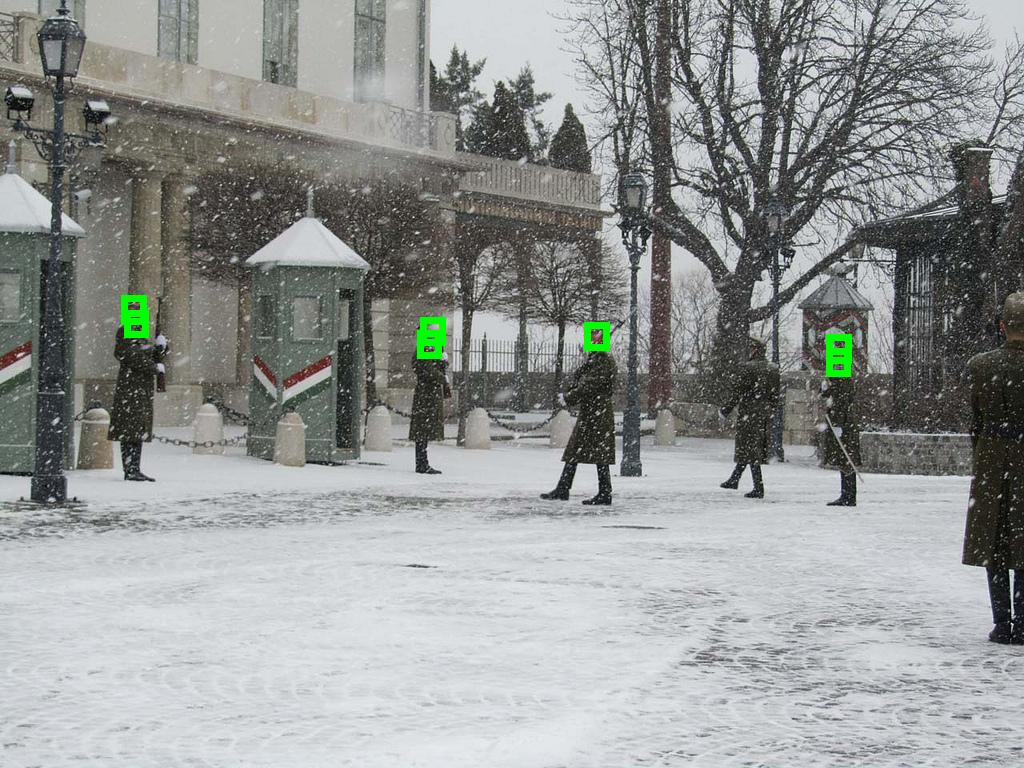}
		\includegraphics[width=0.13\linewidth]{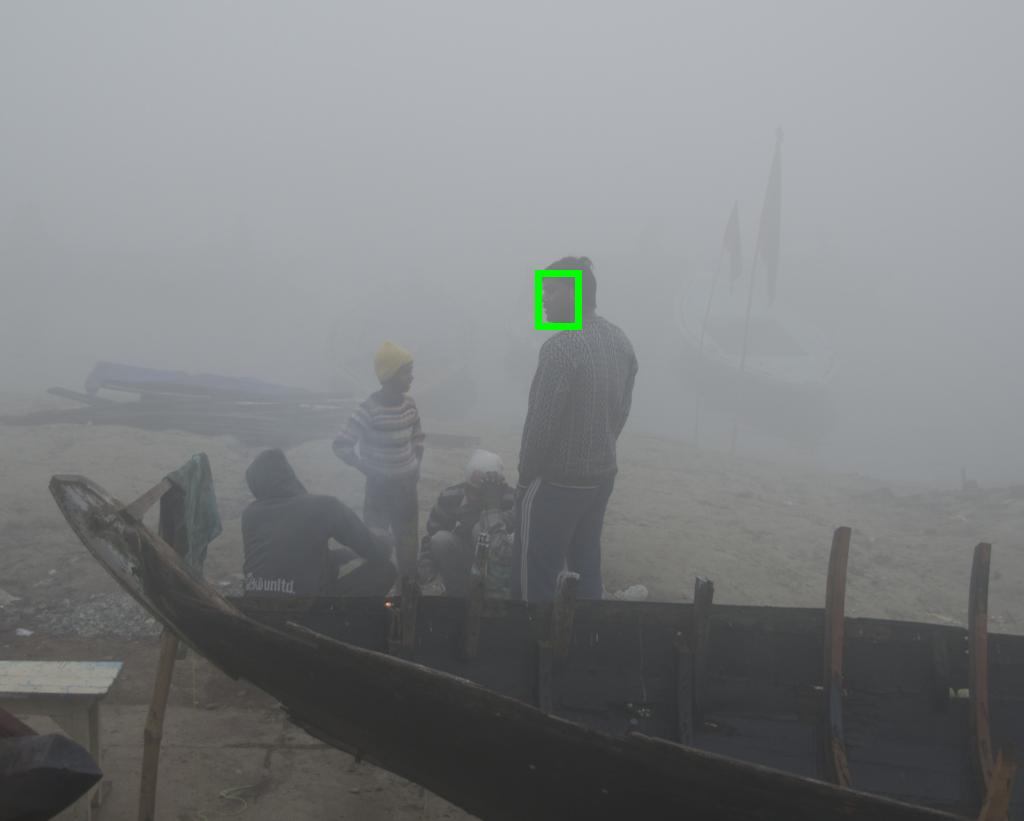}
		\includegraphics[width=0.13\linewidth]{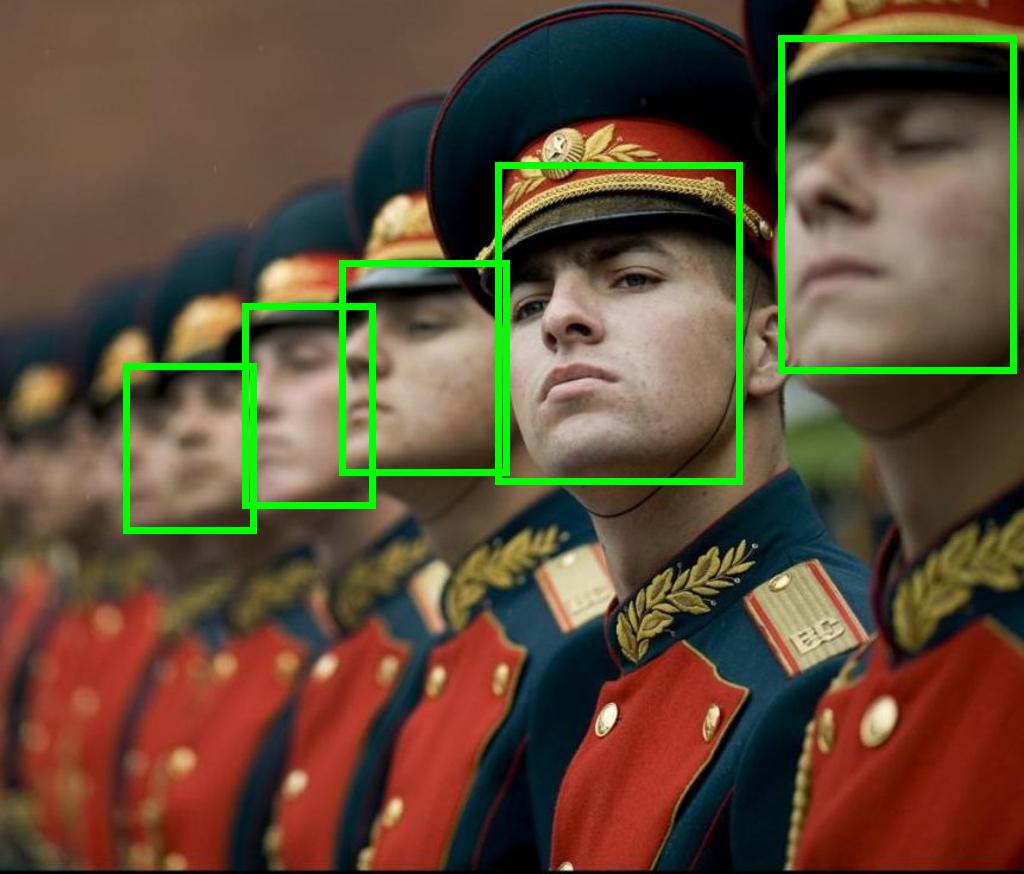}
		\includegraphics[width=0.13\linewidth]{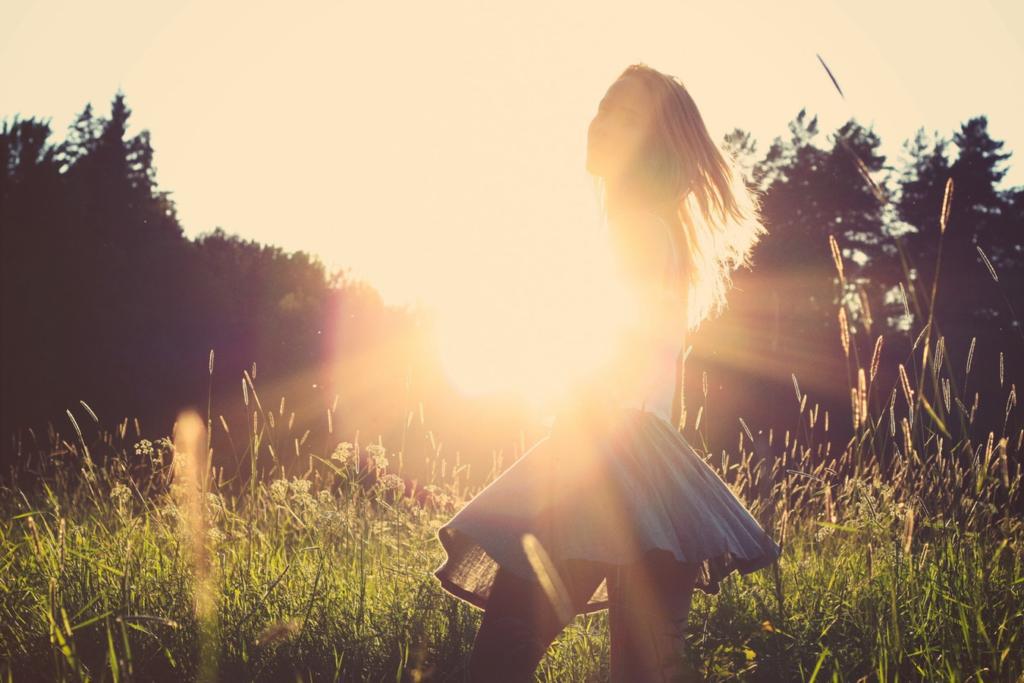}
		\includegraphics[width=0.13\linewidth]{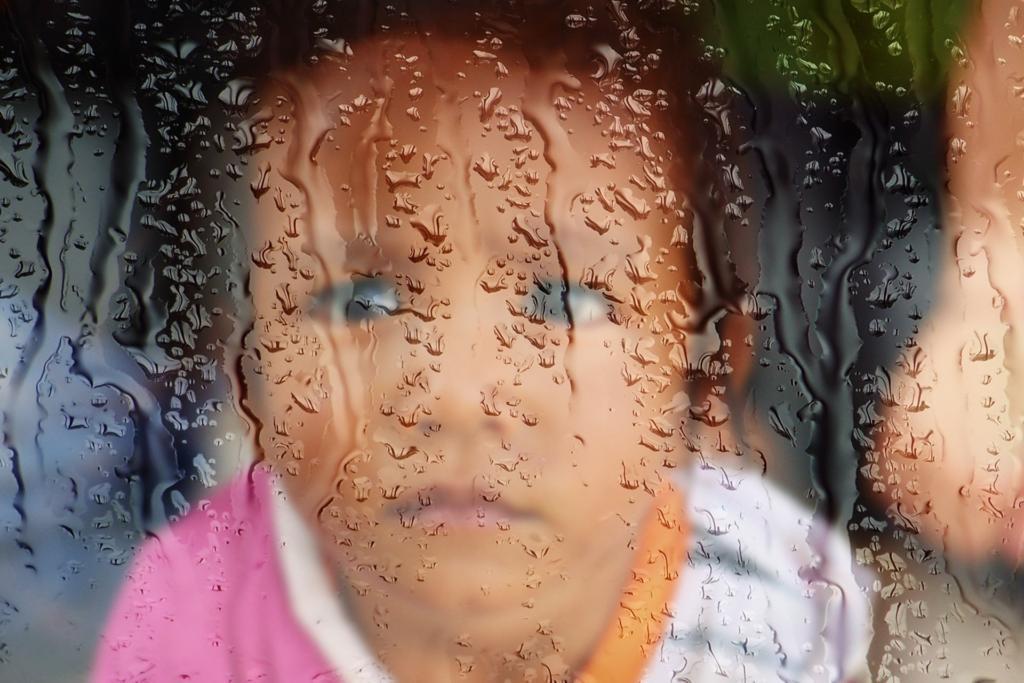}
		\includegraphics[width=0.13\linewidth]{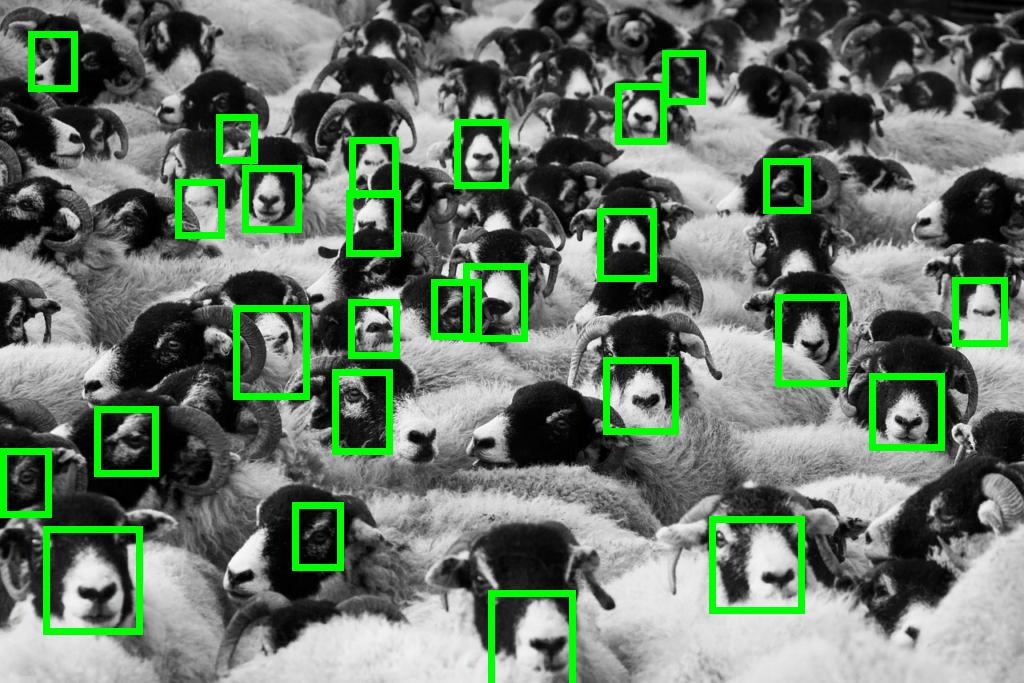}

		\rotatebox{90}{SSH}
		\includegraphics[width=0.13\linewidth]{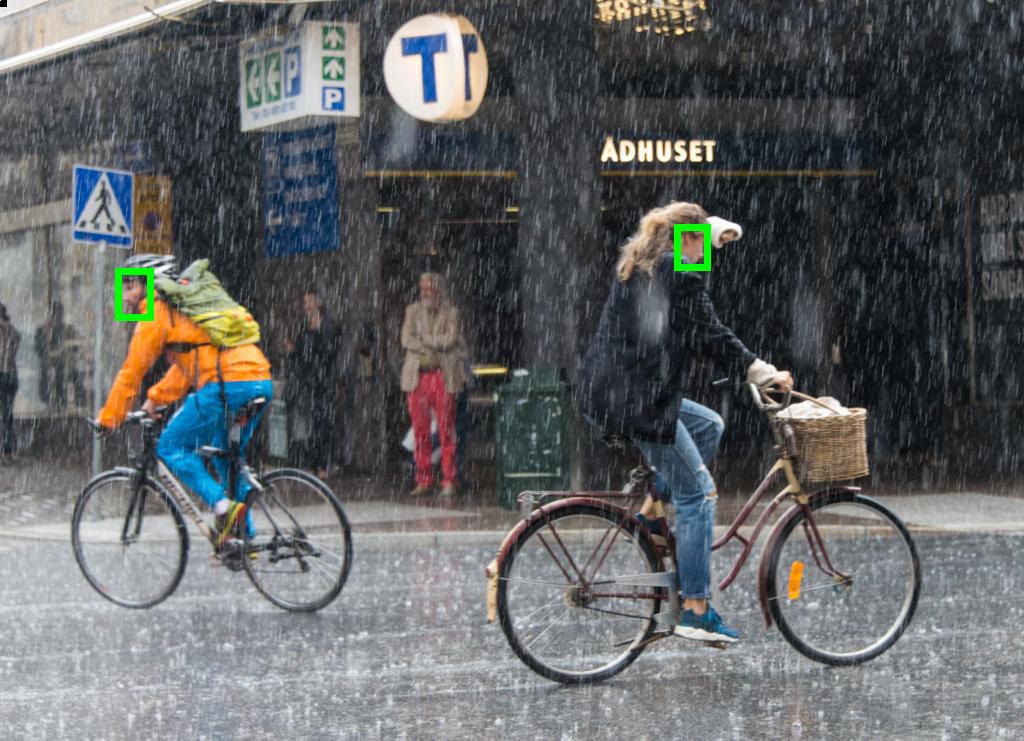}
		\includegraphics[width=0.13\linewidth]{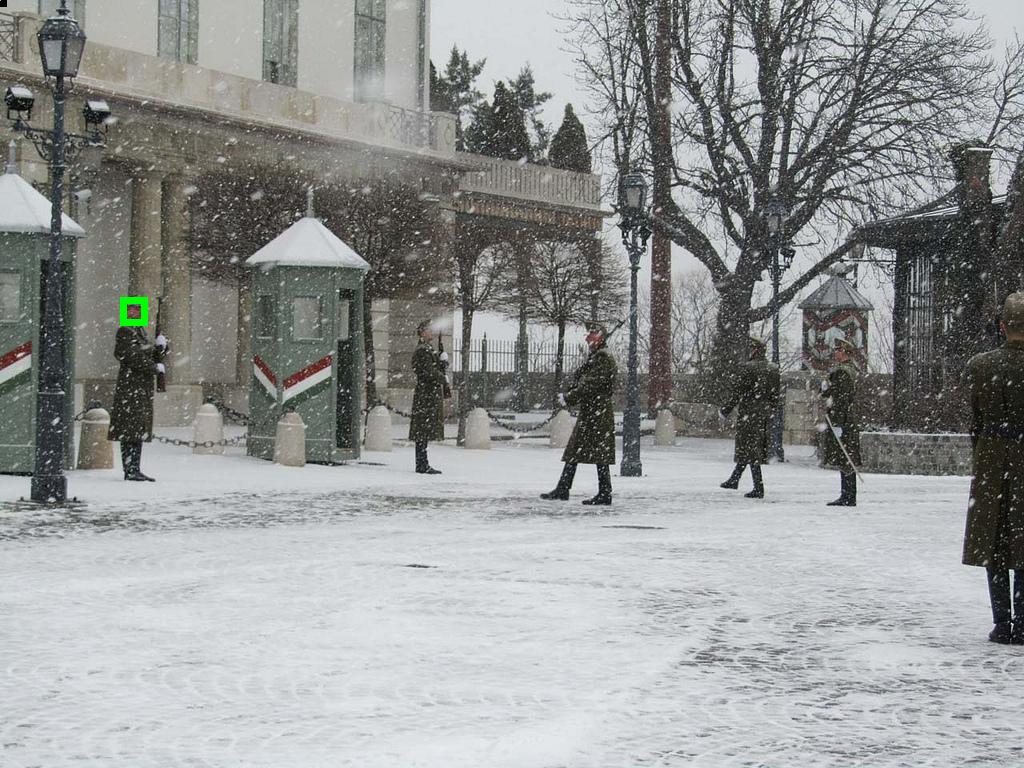}
		\includegraphics[width=0.13\linewidth]{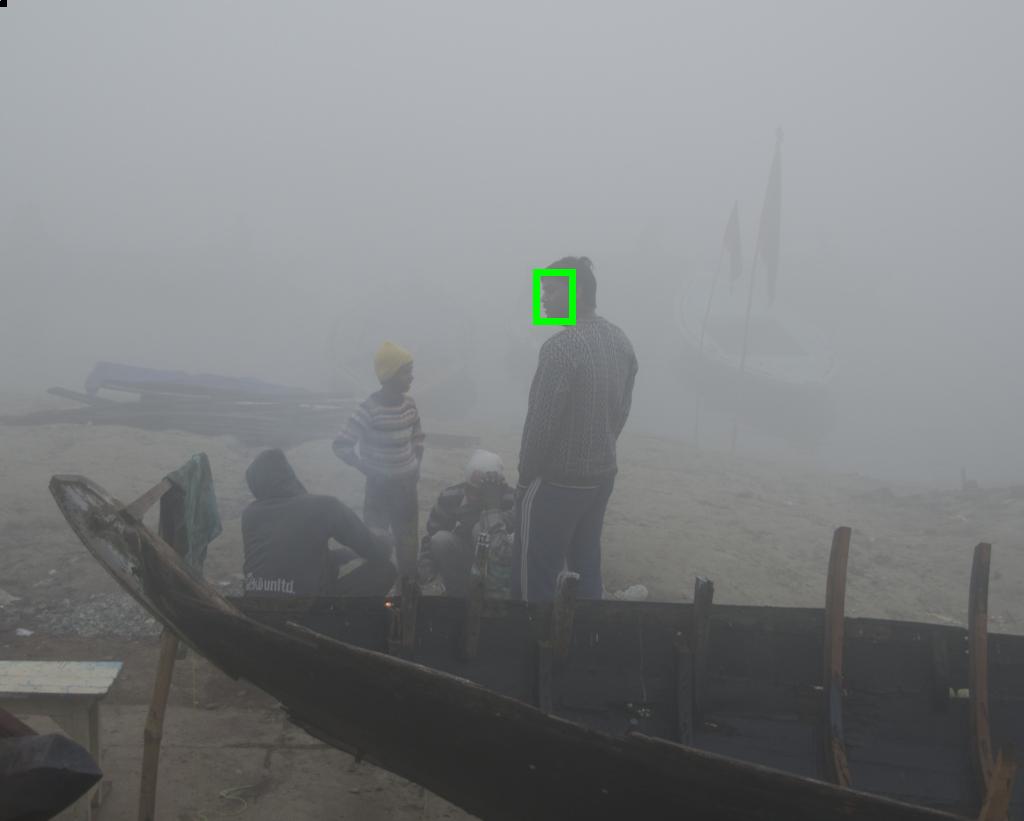}
		\includegraphics[width=0.13\linewidth]{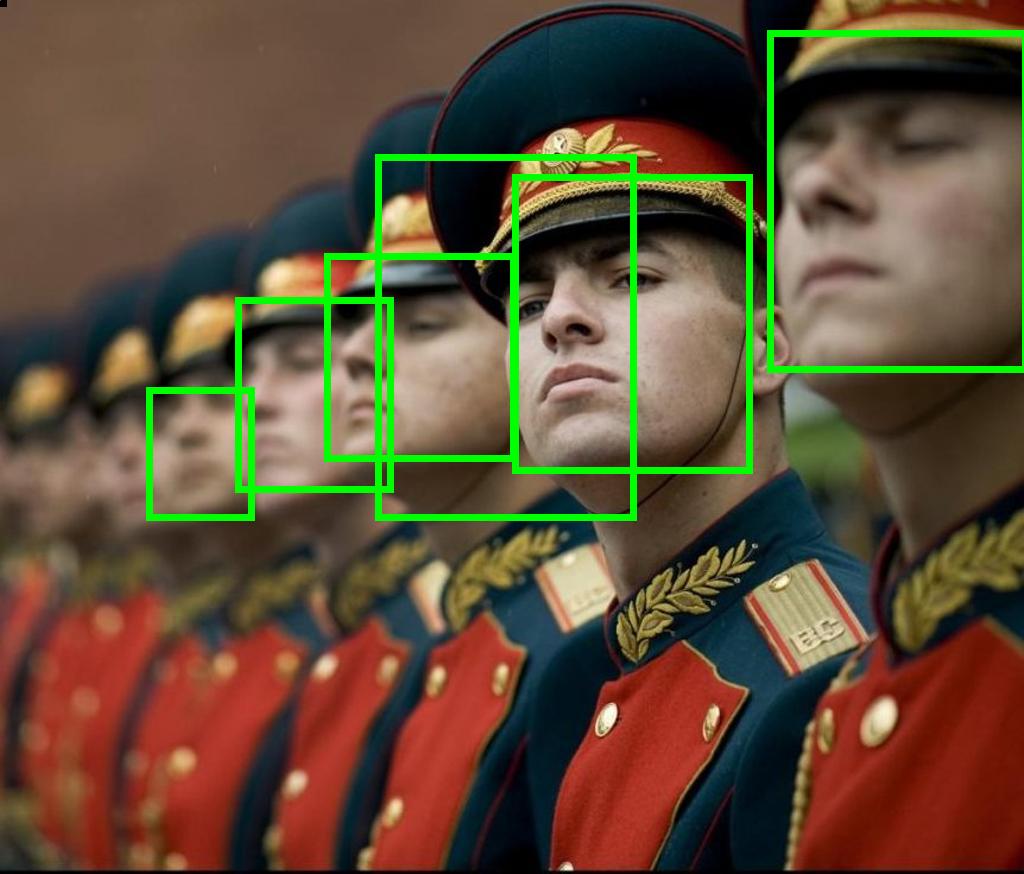}
		\includegraphics[width=0.13\linewidth]{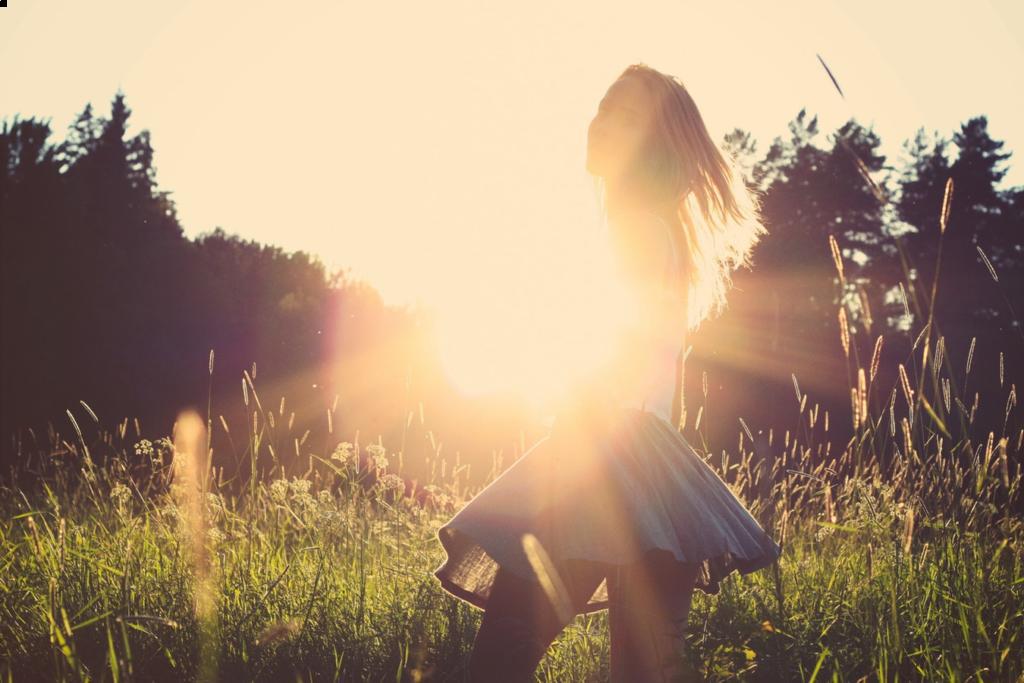}
		\includegraphics[width=0.13\linewidth]{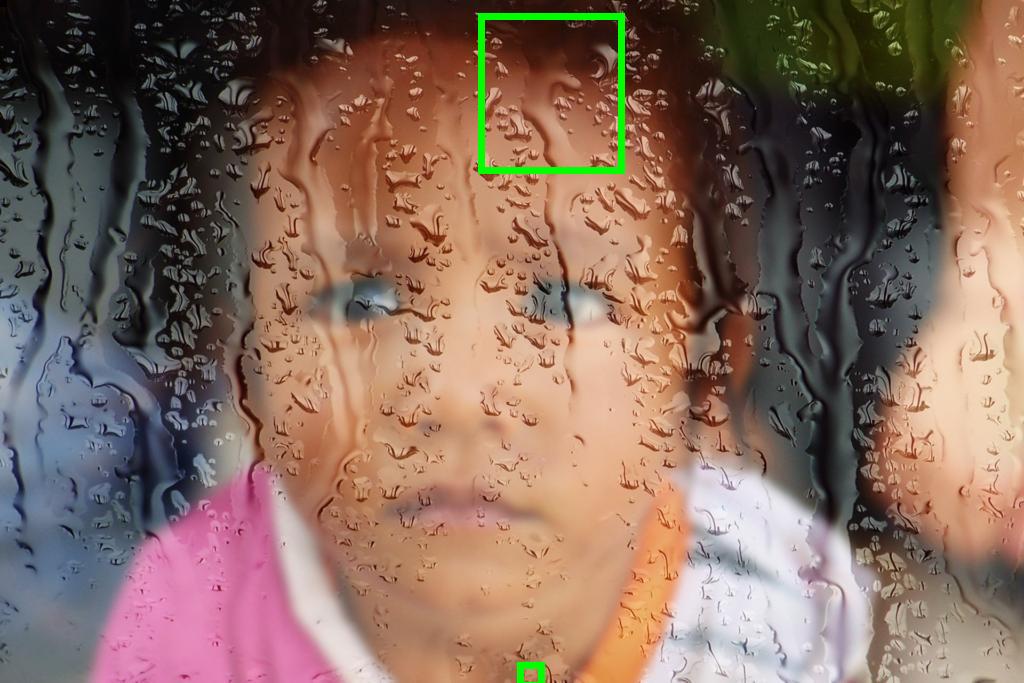}
		\includegraphics[width=0.13\linewidth]{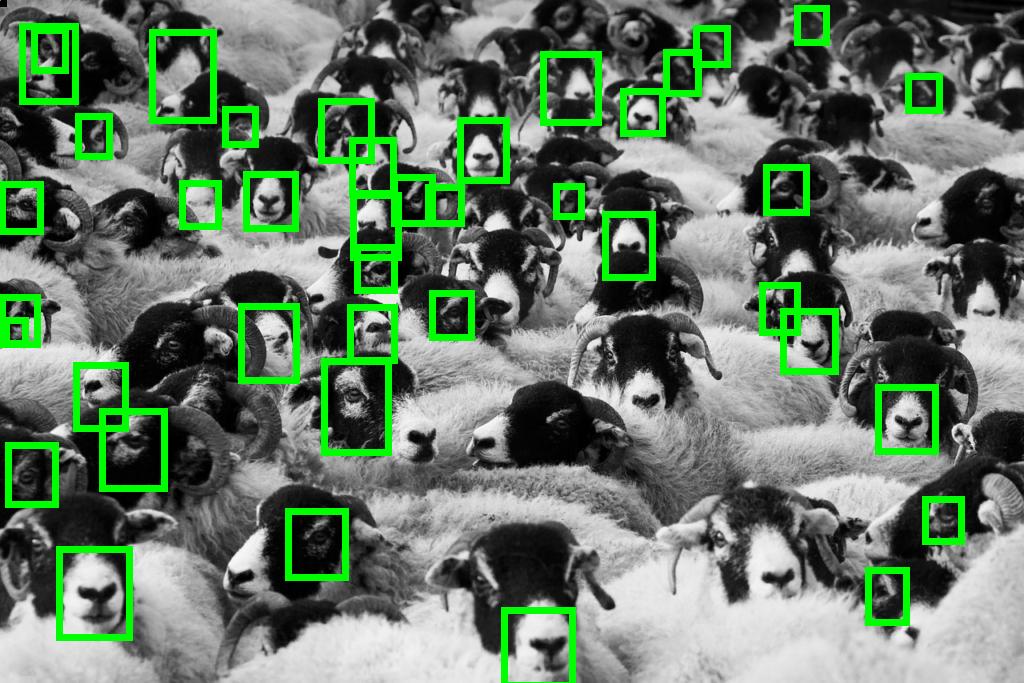}

		\rotatebox{90}{S3FD}
		\includegraphics[width=0.13\linewidth]{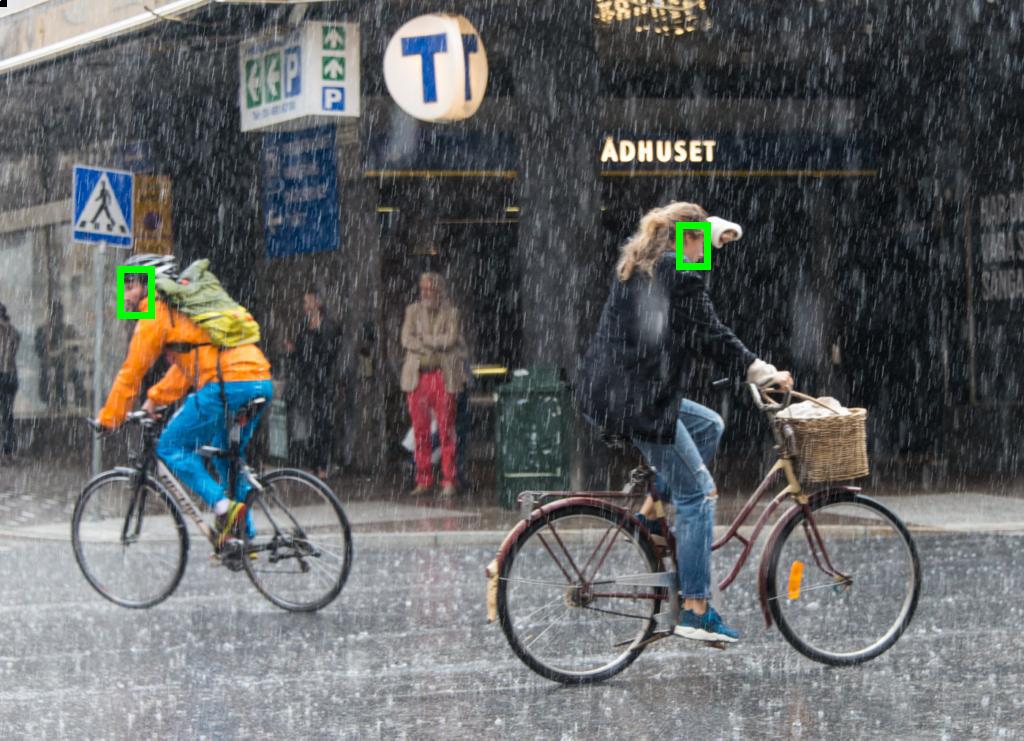}
		\includegraphics[width=0.13\linewidth]{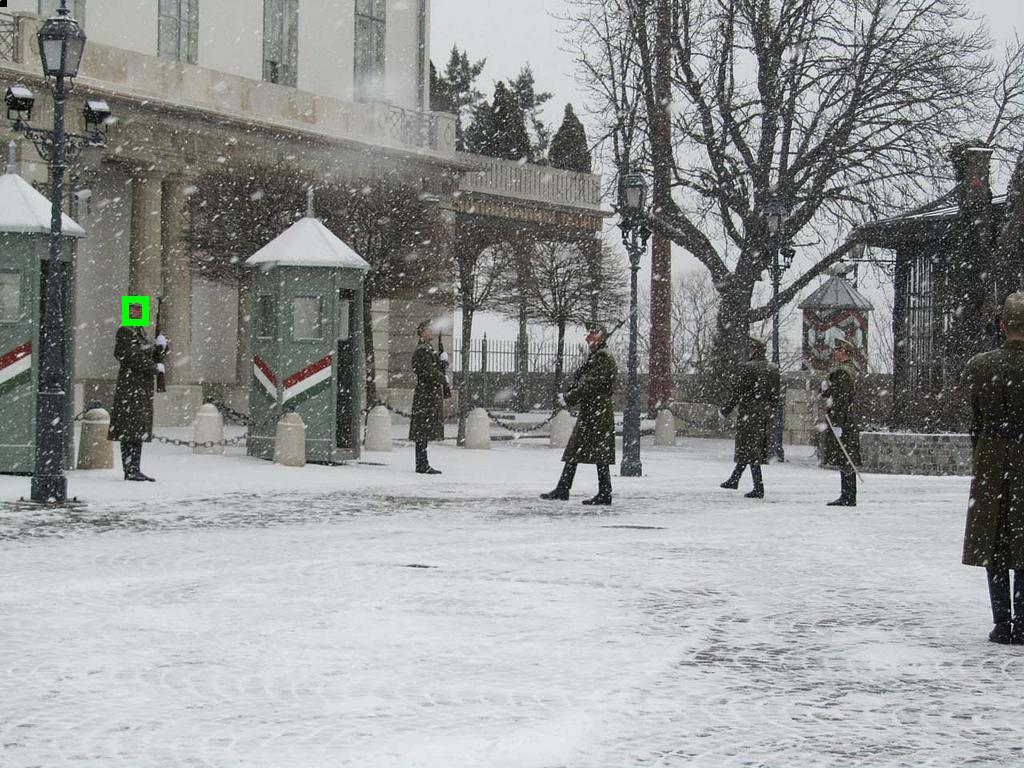}
		\includegraphics[width=0.13\linewidth]{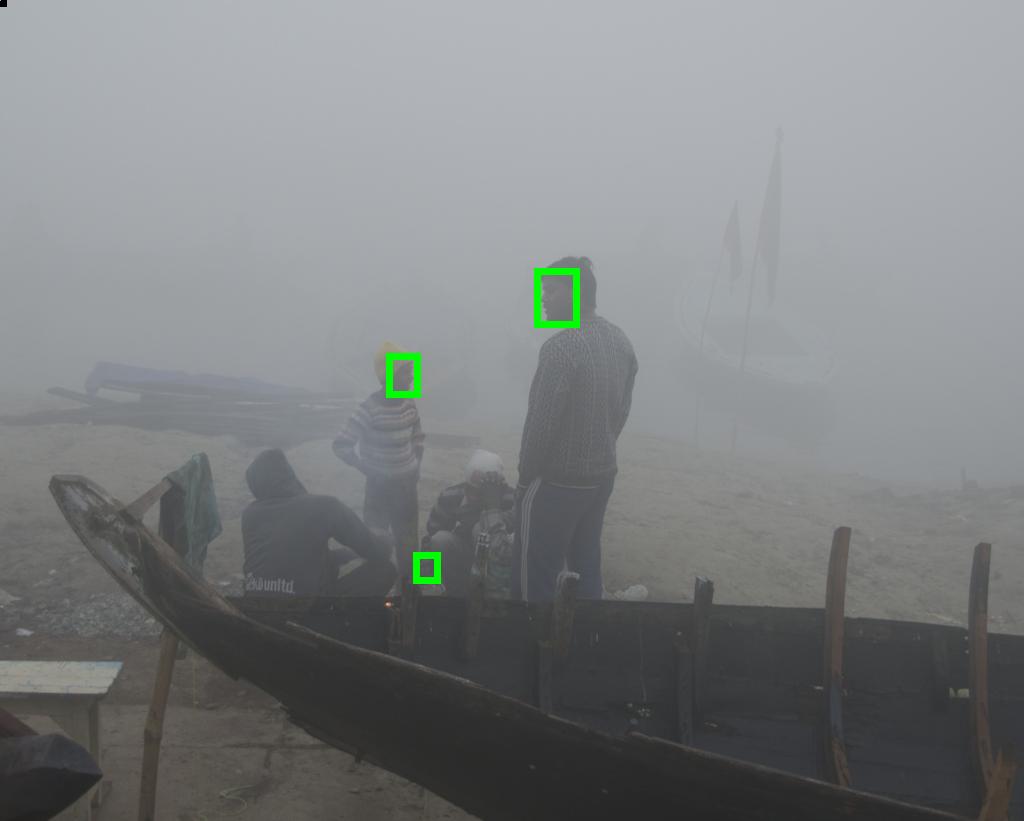}
		\includegraphics[width=0.13\linewidth]{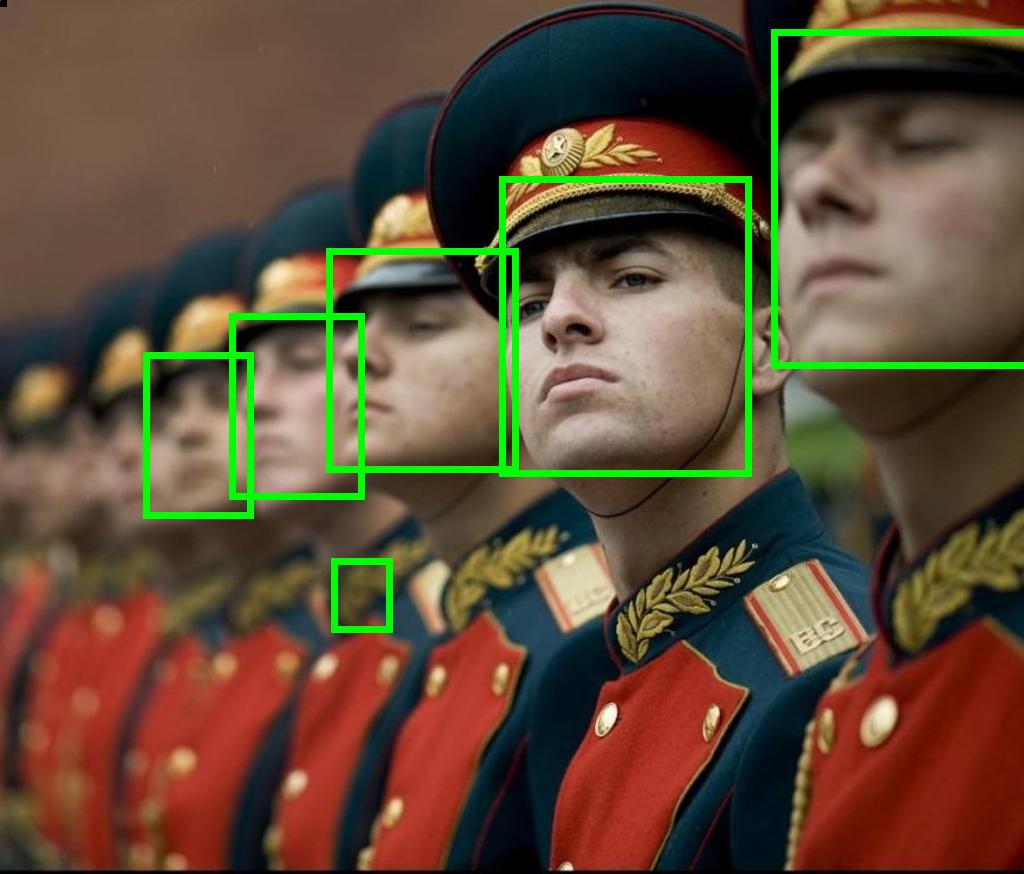}
		\includegraphics[width=0.13\linewidth]{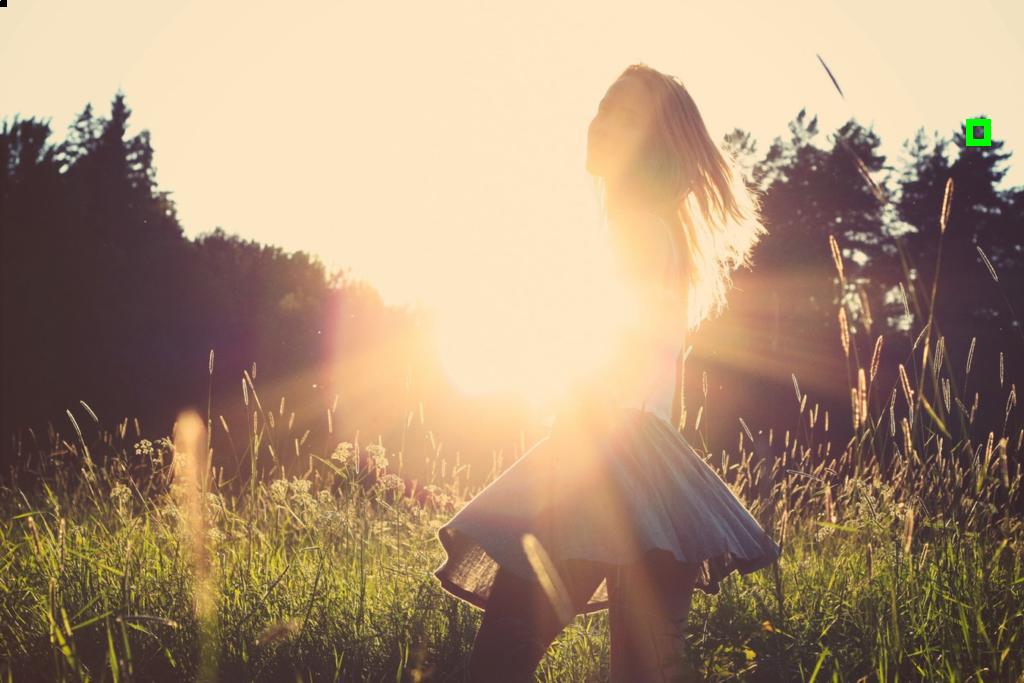}
		\includegraphics[width=0.13\linewidth]{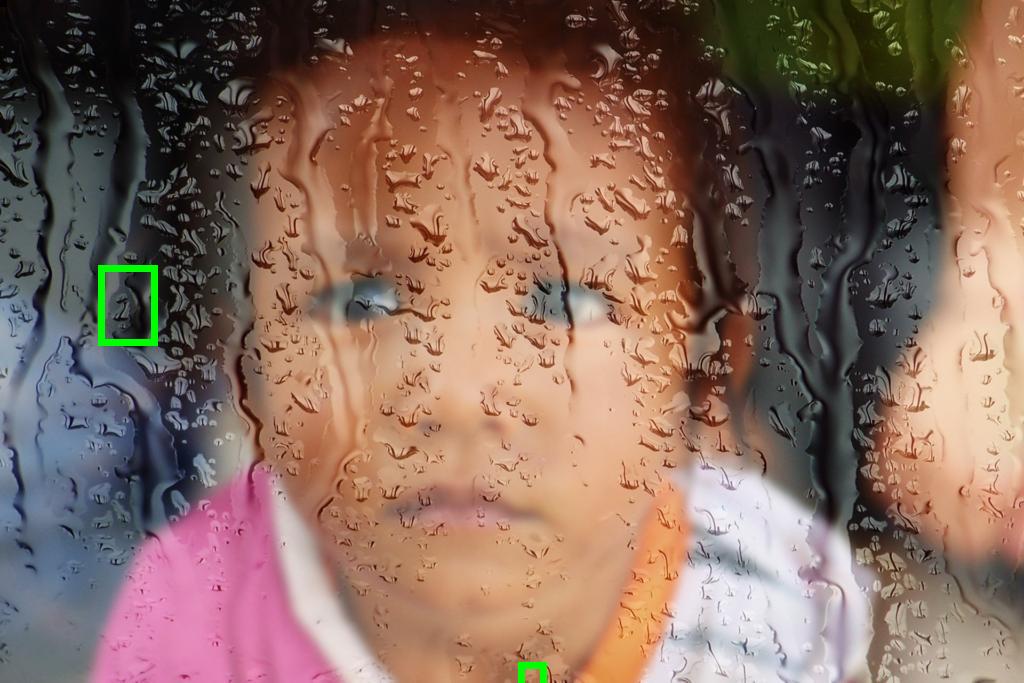}
		\includegraphics[width=0.13\linewidth]{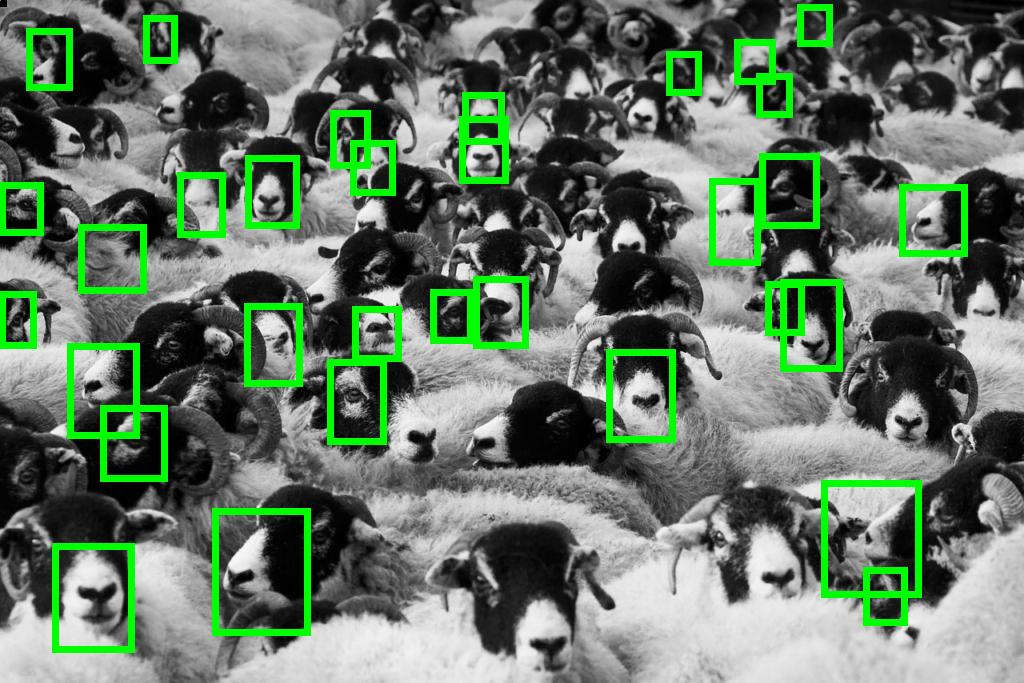}

		\rotatebox{90}{HR-ER}
		\includegraphics[width=0.13\linewidth]{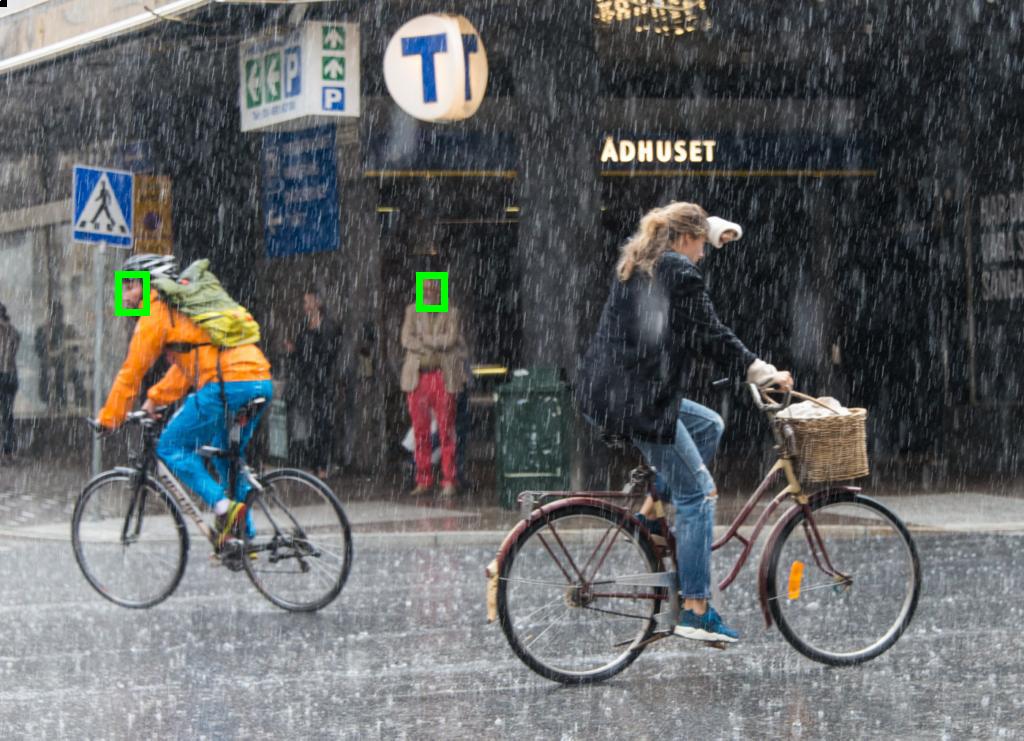}
		\includegraphics[width=0.13\linewidth]{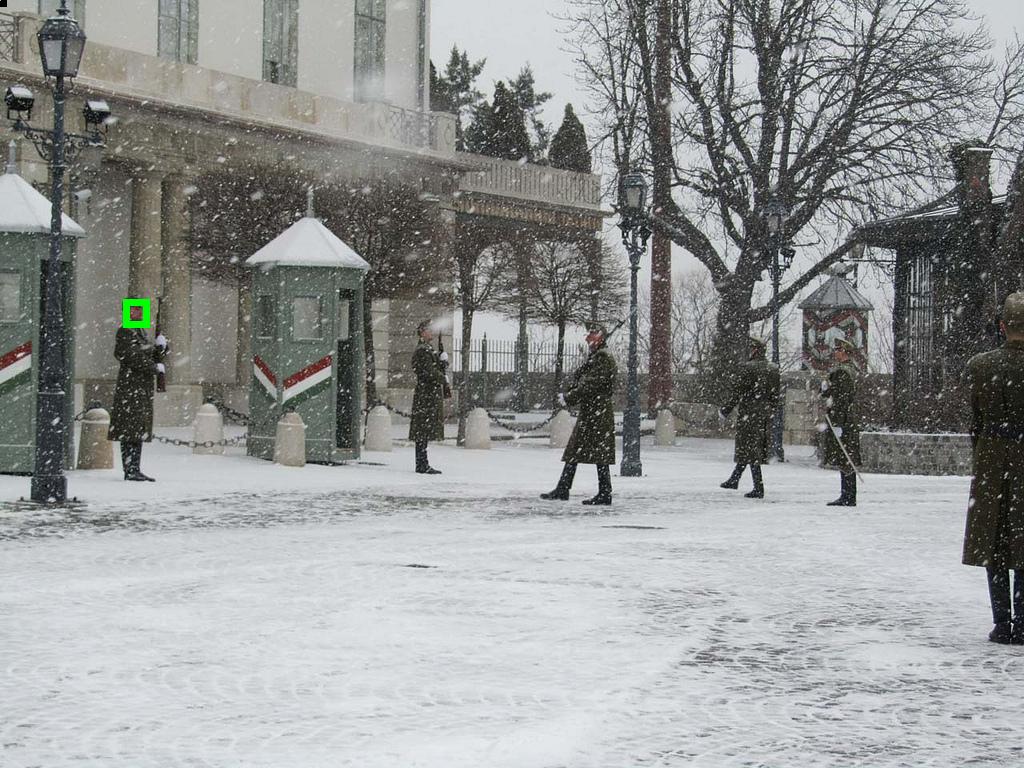}
		\includegraphics[width=0.13\linewidth]{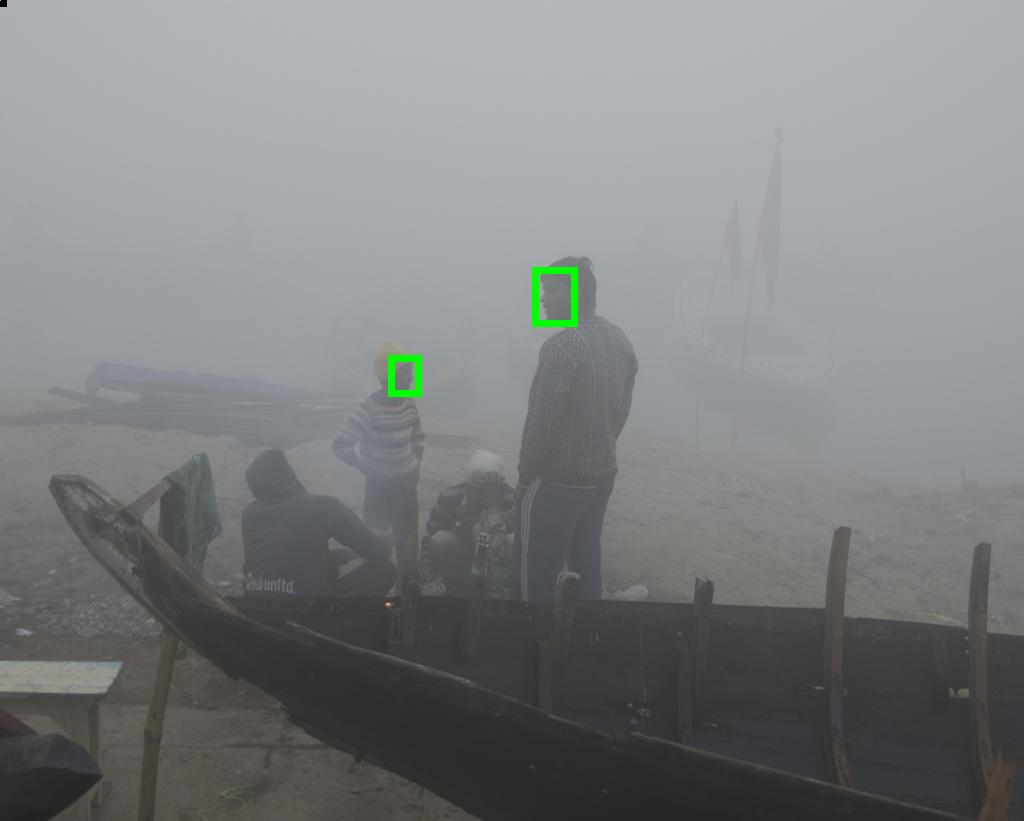}
		\includegraphics[width=0.13\linewidth]{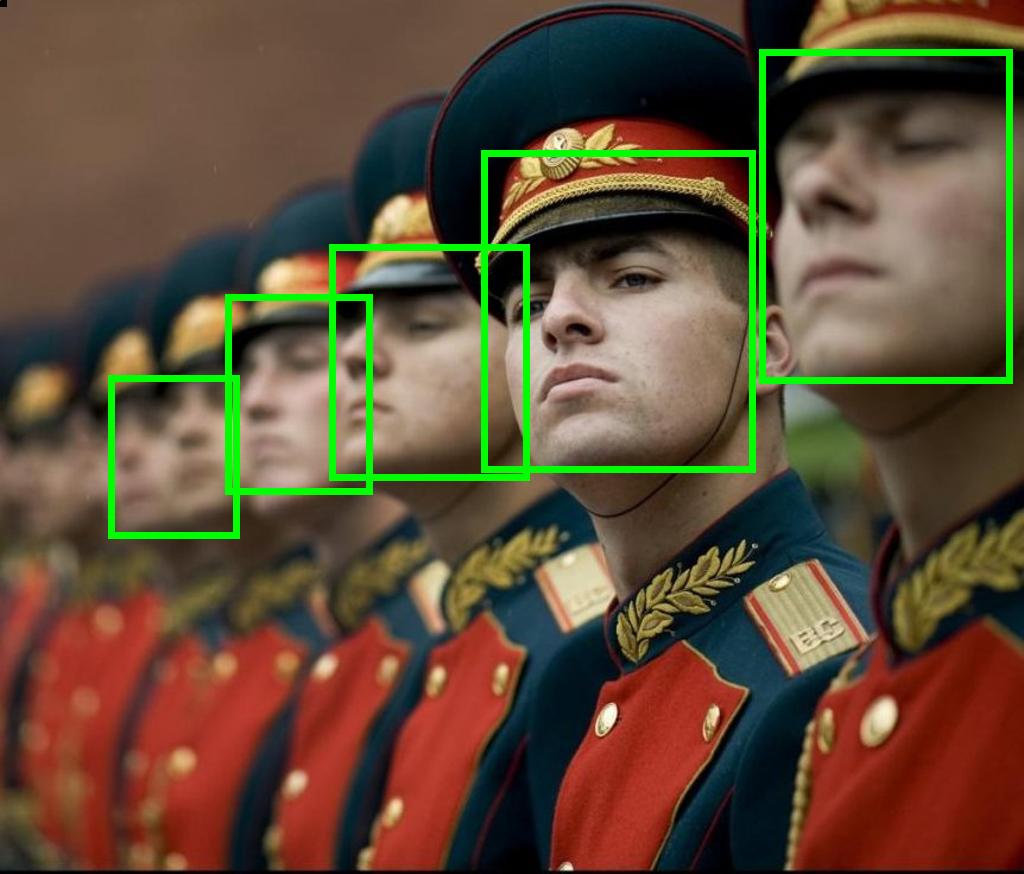}
		\includegraphics[width=0.13\linewidth]{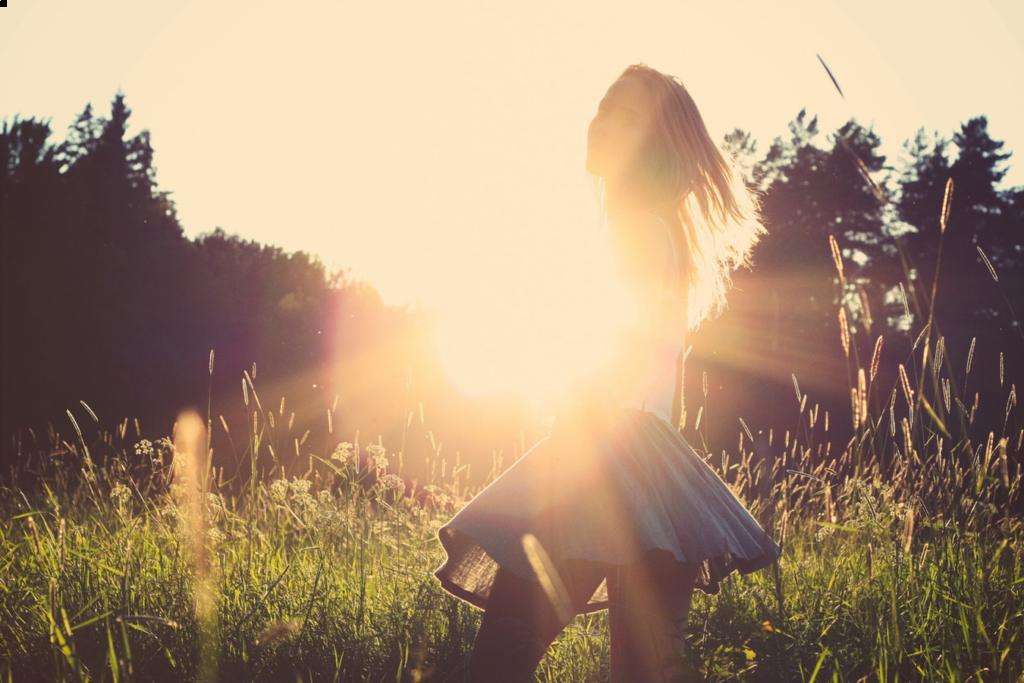}
		\includegraphics[width=0.13\linewidth]{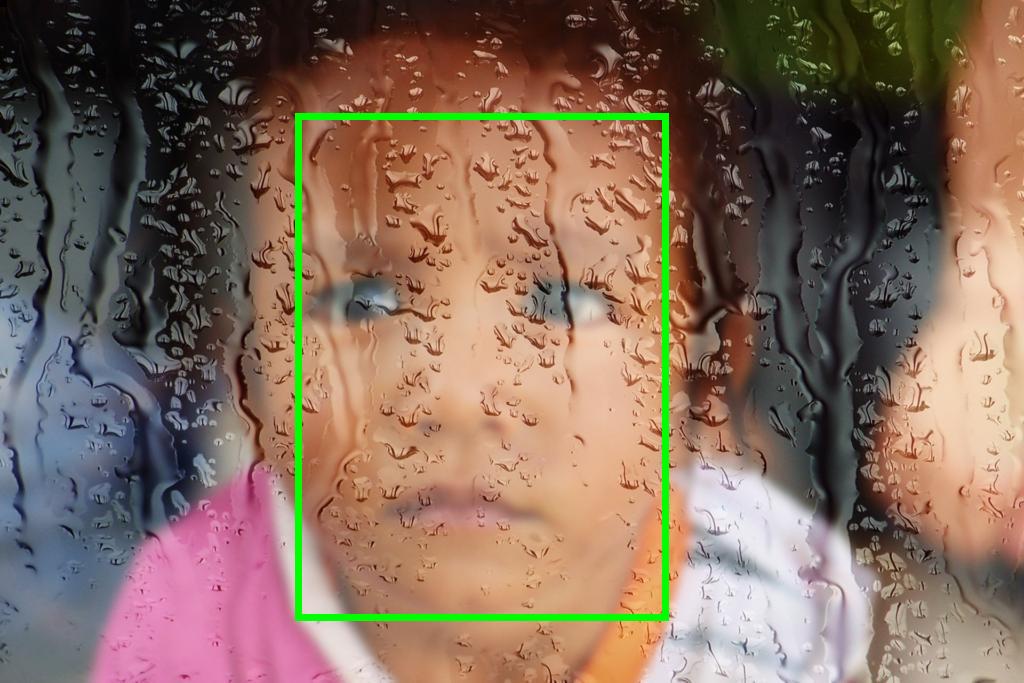}
		\includegraphics[width=0.13\linewidth]{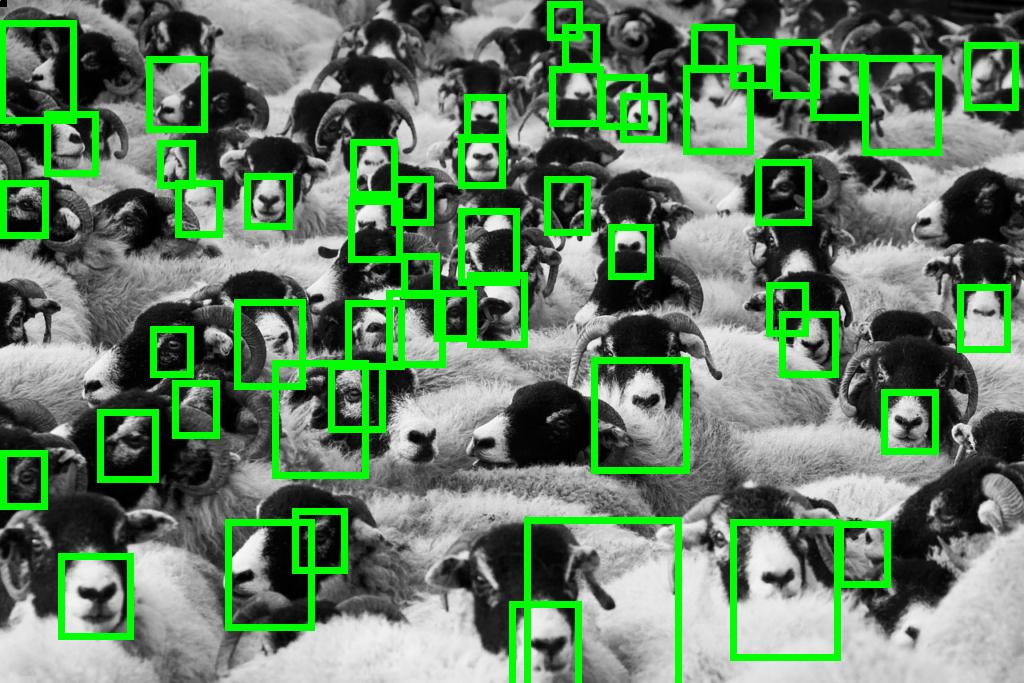} \\

\hskip 15pt Rain \hskip 45pt Snow \hskip 45pt Haze \hskip 45pt Blur \hskip 30pt Illumination \hskip 10pt Lens impediments \hskip 10pt Distractors \hskip 15pt 
	\end{center}
	\vskip -14pt \caption{Sample face detection results on the proposed UFDD dataset.}
	\label{fig:fail_detect}
\end{figure*}

\begin{figure*}[t]
	\centering
	\begin{minipage}{.3\textwidth}
		\centering
		\includegraphics[width=.8\linewidth, height=0.7\linewidth]{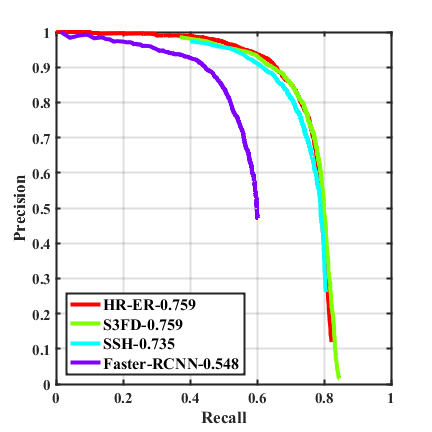}
		\captionsetup{labelformat=empty}
		\captionsetup{justification=centering}
	    \vskip -8pt
		\caption*{Rain}
	\end{minipage}
	\begin{minipage}{.3\textwidth}
		\centering
		\includegraphics[width=.8\linewidth, height=0.7\linewidth]{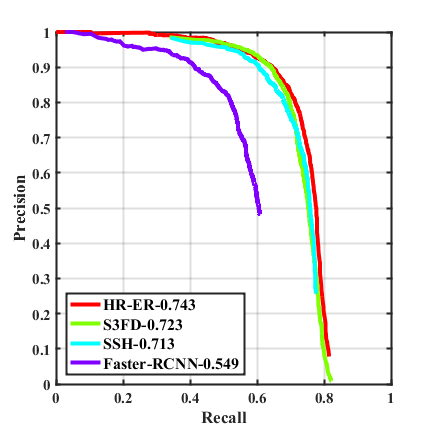}
		\captionsetup{labelformat=empty}
		\captionsetup{justification=centering}
	    \vskip -8pt
		\caption*{Snow}
	\end{minipage}
	\begin{minipage}{.3\textwidth}
		\centering
		\includegraphics[width=.8\linewidth, height=0.7\linewidth]{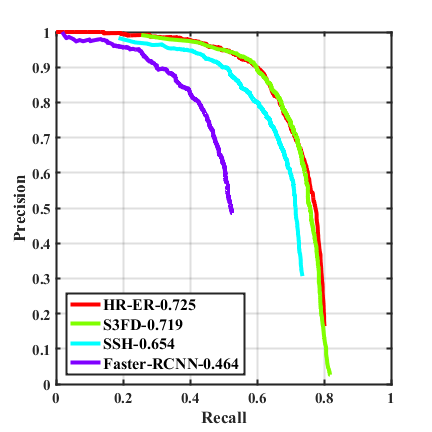}
		\captionsetup{labelformat=empty}
		\captionsetup{justification=centering}
	    \vskip -8pt
		\caption*{Haze}
	\end{minipage}
	\begin{minipage}{.3\textwidth}
		\centering
		\includegraphics[width=.8\linewidth, height=0.7\linewidth]{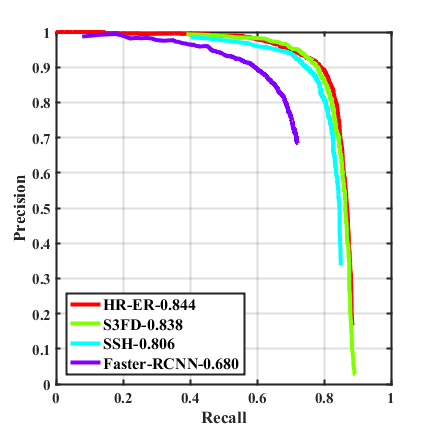}
		\captionsetup{labelformat=empty}
		\captionsetup{justification=centering}
	    \vskip -8pt
		\caption*{Blur}
	\end{minipage}
	\begin{minipage}{.3\textwidth}
		\centering
		\includegraphics[width=.8\linewidth, height=0.7\linewidth]{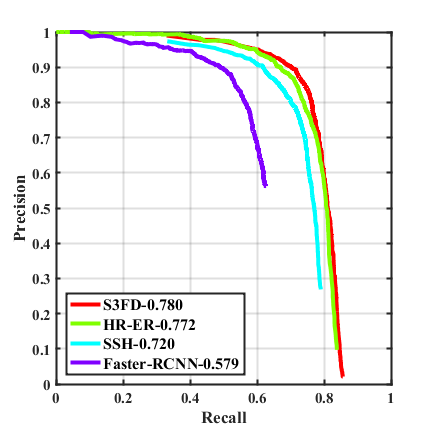}
		\captionsetup{labelformat=empty}
		\captionsetup{justification=centering}
	    \vskip -8pt
		\caption*{Illumination}
	\end{minipage}
	\begin{minipage}{.3\textwidth}
		\centering
		\includegraphics[width=.8\linewidth, height=0.7\linewidth]{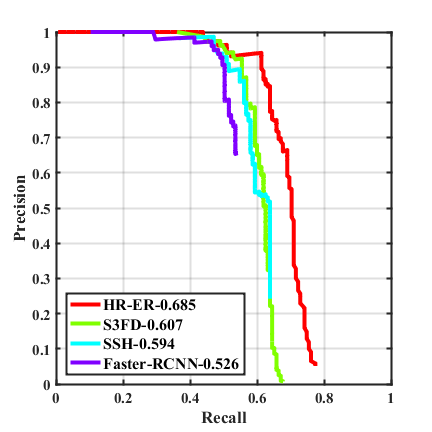}
		\captionsetup{labelformat=empty}
		\captionsetup{justification=centering}
	    \vskip -8pt
		\caption*{Lens impediments}
	\end{minipage}	
	\vskip -10pt\caption{Cohort Analysis: Individual precision-recall curves of different face detection algorithms on the proposed UFDD dataset. Note that the face detectors are pre-trained on the WIDER FACE dataset.} \label{fig:over}
	\vskip -18pt
	\label{fig:prcurves_cohort}
\end{figure*}

\subsection{Cohort Analysis}
In this section, we individually analyze the effect of different conditions such as rain, haze, \etc on the performance of recent state-of-the-art face detection methods\footnote{All four methods use WIDER FACE as the source training set and these pre-trained models are evaluated on the UFDD dataset.}. Results of this study (precision-recall curves) are presented in Fig. \ref{fig:prcurves_cohort}. Detection results on a sample image for all the four benchmark methods are shown in Fig. \ref{fig:fail_detect}. It can be clearly observed from these figures that all the degradations hinder the performance of the recent state-of-the-art detectors. These degradations introduce different kinds of artifacts in the feature maps, thereby resulting in a slightly modified representation as compared to the original representation. Since the existing methods are trained on the datasets that do not necessarily contain large number of images with these conditions, such methods do not generalize well to new conditions.

Performance drops are observed under all degradations, although to different degrees. Among all degradations, the presence of  haze and lens impediments have a relatively more impact, which is probably because these conditions severely degrade the image and the problem is further aggravated due to the fact that the WIDER FACE dataset does not contain many hazy and lens impediments images. A surprising observation is that the HR-ER method \cite{hu2017finding} consistently performs better than more recent methods such as SSH \cite{najibi2017ssh} and S3FD \cite{zhang2017s3fd}. This is especially important considering the fact that SSH and S3FD perform better on the WIDER FACE dataset as compared to HR-ER. Based on this observation, we may conclude that HR-ER has better generalization ability as compared to the other detectors. In the following, we  discuss the results for  each condition in detail.

\begin{table}[t!]
	\centering
	\caption{The mAP scores corresponding to different detectors on the UFDD dataset with and without distractors and their differences.}
	\label{tab:map_dist}
	\vskip -8pt
	\resizebox{0.8\linewidth}{!}{%
		\begin{tabular}{|l|c|c|c|}
			\hline
			DB & UFDD & UFDD without distractors & Difference\\ \hline
			Faster-RCNN \cite{ren2015faster} & 0.521               & 0.564   & 0.043       \\ \hline
			SSH    \cite{najibi2017ssh}     & 0.695               & 0.725   & 0.030       \\ \hline
			S3FD  \cite{najibi2017ssh}      & 0.725               & 0.761   & 0.036       \\ \hline
			HR-ER  \cite{hu2017finding}     & 0.742               & 0.767   & 0.025       \\ \hline
		\end{tabular}
	}
\end{table}

\noindent \textbf{Rain:} The presence of rain streaks alters the high frequency components in an image, thus changing the filter responses. This results in degradation of visual quality and poor detection performance \cite{zhang2017rain}. The problem is further exacerbated when images containing occluded faces are degraded with rain streaks. 

\noindent \textbf{Snow:} Similar to rain, the presence of snow also degrades the performance of face detection since it blocks certain parts of the face (as shown in Fig. \ref{fig:fail_detect}. However, the degradation observed due to snow is comparatively higher which could be due to the fact that the presence of snow results in larger degrees of occlusion as compared to that caused by rain. 

\noindent \textbf{Haze:} Haze, caused by the absorption or reflection of light by floating particles in the air, results in low image contrast affecting the visibility of faces in images. In addition to causing serious degradation of image quality, haze causes a significant drop in face detection performance. As shown in the third column in Fig. \ref{fig:fail_detect}, the faces are less visible and tend to be darker due to the presence of haze. It can be observed that haze  causes relatively more degradation in the performance compared to rain and snow. 

\noindent \textbf{Blur:} Blur, caused either by camera shake or due to depth, results in loss of crucial high frequency details in an image. This loss of information results in considerable difficulties for face detection. Since existing face detectors are trained on datasets containing sharp and high-quality images, the representations learned by these detectors are not robust to blurry images.  

\noindent \textbf{Illumination:} Extreme illumination conditions such as  excessive brightness or darkness affects the visibility of faces.  It can be observed from  Fig. \ref{fig:fail_detect} that all four methods are unable to detect the faces in images with extreme illumination conditions. 

\noindent \textbf{Lens impediments:} Lens impediments, caused by the presence of dirt particles or water droplets on the camera lens, introduces sudden discontinuities in frequencies and hence, large variations in focus in the captured images. As shown in the last column in Fig. \ref{fig:fail_detect}, the presence of water droplets results in regions in the image that have different focus.  This results either in false detections or miss-detections. 

\begin{figure}[t!]
	\begin{center}				
		\includegraphics[width=.5\linewidth, height=0.5\linewidth]{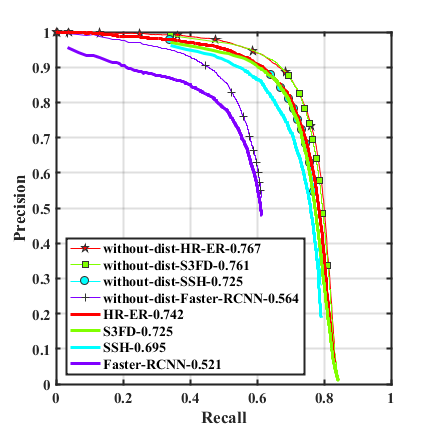}	
	\end{center}
	\vskip -20pt \caption{Evaluation results of different algorithms on the proposed UFDD dataset with and without distractors.} 
	\label{fig:distractors}
\end{figure}

\noindent \textbf{Distractors:} Distractors are images that do not contain human faces. Example of distractor images are the ones containing hand regions, animal faces, \etc These images contain regions which can be easily confused as faces and hence, these kind of images result in high false positive rate. It can be observed from  Table \ref{tab:map_dist} and Fig.~\ref{fig:distractors} that  the detection accuracies drop drastically in the presence of distractor images. Similar observation can be made from the last column in Fig. \ref{fig:fail_detect}.

\section{Conclusion}
\label{sec:conclusion}
We identified the next set of challenges that plague the face detection task. While existing datasets capture large variations in different factors, these newly identified conditions are largely ignored. To overcome this, we collected a new UFDD dataset  that specifically captures different image degradations due to weather conditions such as rain, snow, haze, \etc and blur-based degradations. In addition, the dataset also consists of several images known as distractors that contain non-human faces and objects. We benchmarked recent state-of-the-art face detection algorithms on this newly proposed dataset and demonstrate a significant gap in their performance.  Additionally, in order to provide an insight for design of future algorithms, we also presented a detailed cohort analysis that studies the effect of different conditions on the detection performance.


\section*{Acknowledgments}
{\footnotesize{This research is based upon work supported by the Office of the Director of National Intelligence (ODNI), Intelligence Advanced Research Projects Activity (IARPA), via IARPA R\&D Contract No. 2014-14071600012. The views and conclusions contained herein are those of the authors and should not be interpreted as necessarily representing the official policies or endorsements, either expressed or implied, of the ODNI, IARPA, or the U.S. Government. The U.S. Government is authorized to reproduce and distribute reprints for Governmental purposes notwithstanding any copyright annotation thereon.}}

{\small
\bibliographystyle{ieee}
\bibliography{egbib}

\begin{thebibliography}{10}\itemsep=-1pt

\bibitem{pyfaster}
https://github.com/playerkk/face-py-faster-rcnn.

\bibitem{hash}
https://pypi.python.org/pypi/imagehash.

\bibitem{synth-lens}
https://www.how2des.com/2014/10/realistic-water-drops-effect-on-a-wet-frosted-glass.html.

\bibitem{matpoisson}
https://www.mathworks.com/help/images/ref/imnoise.html.

\bibitem{synth-rain}
https://www.photoshopessentials.com/photo-effects/photoshop-weather-effects-rain/.

\bibitem{synth-snow}
https://www.photoshopessentials.com/photo-effects/photoshop-weather-effects-snow/.

\bibitem{brubaker2008design}
S.~C. Brubaker, J.~Wu, J.~Sun, M.~D. Mullin, and J.~M. Rehg.
\newblock On the design of cascades of boosted ensembles for face detection.
\newblock {\em International Journal of Computer Vision}, 77(1-3):65--86, 2008.

\bibitem{cai2016unified}
Z.~Cai, Q.~Fan, R.~S. Feris, and N.~Vasconcelos.
\newblock A unified multi-scale deep convolutional neural network for fast
  object detection.
\newblock In {\em European Conference on Computer Vision}, pages 354--370.
  Springer, 2016.

\bibitem{chen2014joint}
D.~Chen, S.~Ren, Y.~Wei, X.~Cao, and J.~Sun.
\newblock Joint cascade face detection and alignment.
\newblock In {\em European Conference on Computer Vision}, pages 109--122.
  Springer, 2014.

\bibitem{di2017gp}
X.~Di, V.~A. Sindagi, and V.~M. Patel.
\newblock Gp-gan: gender preserving gan for synthesizing faces from landmarks.
\newblock {\em arXiv preprint arXiv:1710.00962}, 2017.

\bibitem{everingham2015pascal}
M.~Everingham, S.~A. Eslami, L.~Van~Gool, C.~K. Williams, J.~Winn, and
  A.~Zisserman.
\newblock The pascal visual object classes challenge: A retrospective.
\newblock {\em International journal of computer vision}, 111(1):98--136, 2015.

\bibitem{gross2010multi}
R.~Gross, I.~Matthews, J.~Cohn, T.~Kanade, and S.~Baker.
\newblock Multi-pie.
\newblock {\em Image and Vision Computing}, 28(5):807--813, 2010.

\bibitem{gunther2017unconstrained}
M.~G{\"u}nther, P.~Hu, C.~Herrmann, C.~H. Chan, M.~Jiang, S.~Yang, A.~R.
  Dhamija, D.~Ramanan, J.~Beyerer, J.~Kittler, et~al.
\newblock Unconstrained face detection and open-set face recognition challenge.
\newblock {\em arXiv preprint arXiv:1708.02337}, 2017.

\bibitem{he2011single}
K.~He, J.~Sun, and X.~Tang.
\newblock Single image haze removal using dark channel prior.
\newblock {\em IEEE transactions on pattern analysis and machine intelligence},
  33(12):2341--2353, 2011.

\bibitem{he2016deep}
K.~He, X.~Zhang, S.~Ren, and J.~Sun.
\newblock Deep residual learning for image recognition.
\newblock In {\em Proceedings of the IEEE conference on computer vision and
  pattern recognition}, pages 770--778, 2016.

\bibitem{hu2017finding}
P.~Hu and D.~Ramanan.
\newblock Finding tiny faces.
\newblock In {\em 2017 IEEE Conference on Computer Vision and Pattern
  Recognition (CVPR)}, pages 1522--1530. IEEE, 2017.

\bibitem{jain2010fddb}
V.~Jain and E.~Learned-Miller.
\newblock Fddb: A benchmark for face detection in unconstrained settings.

\bibitem{jiang2017face}
H.~Jiang and E.~Learned-Miller.
\newblock Face detection with the faster r-cnn.
\newblock In {\em Automatic Face \& Gesture Recognition (FG 2017), 2017 12th
  IEEE International Conference on}, pages 650--657. IEEE, 2017.

\bibitem{klare2015pushing}
B.~F. Klare, B.~Klein, E.~Taborsky, A.~Blanton, J.~Cheney, K.~Allen,
  P.~Grother, A.~Mah, and A.~K. Jain.
\newblock Pushing the frontiers of unconstrained face detection and
  recognition: Iarpa janus benchmark a.
\newblock In {\em Proceedings of the IEEE Conference on Computer Vision and
  Pattern Recognition}, pages 1931--1939, 2015.

\bibitem{koestinger2011annotated}
M.~Koestinger, P.~Wohlhart, P.~M. Roth, and H.~Bischof.
\newblock Annotated facial landmarks in the wild: A large-scale, real-world
  database for facial landmark localization.
\newblock In {\em Computer Vision Workshops (ICCV Workshops), 2011 IEEE
  International Conference on}, pages 2144--2151. IEEE, 2011.

\bibitem{noise-iso}
M.~Levoy.
\newblock Stanford lecture notes, digital photography, 'noise and iso', 2014.
\newblock
  https://graphics.stanford.edu/courses/cs448a-10/sensors-noise-14jan10-opt.pdf.

\bibitem{li2014efficient}
H.~Li, Z.~Lin, J.~Brandt, X.~Shen, and G.~Hua.
\newblock Efficient boosted exemplar-based face detection.
\newblock In {\em Proceedings of the IEEE Conference on Computer Vision and
  Pattern Recognition}, pages 1843--1850, 2014.

\bibitem{li2015convolutional}
H.~Li, Z.~Lin, X.~Shen, J.~Brandt, and G.~Hua.
\newblock A convolutional neural network cascade for face detection.
\newblock In {\em Proceedings of the IEEE Conference on Computer Vision and
  Pattern Recognition}, pages 5325--5334, 2015.

\bibitem{liu2016ssd}
W.~Liu, D.~Anguelov, D.~Erhan, C.~Szegedy, S.~Reed, C.-Y. Fu, and A.~C. Berg.
\newblock Ssd: Single shot multibox detector.
\newblock In {\em European conference on computer vision}, pages 21--37.
  Springer, 2016.

\bibitem{mathias2014face}
M.~Mathias, R.~Benenson, M.~Pedersoli, and L.~Van~Gool.
\newblock Face detection without bells and whistles.
\newblock In {\em European Conference on Computer Vision}, pages 720--735.
  Springer, 2014.

\bibitem{mazeiarpa-IJB-C}
B.~Maze, J.~Adams, J.~A. Duncan, N.~Kalka, T.~Miller, C.~Otto, A.~K. Jain,
  W.~T. Niggel, J.~Anderson, J.~Cheney, et~al.
\newblock Iarpa janus benchmark--c: Face dataset and protocol.

\bibitem{najibi2017ssh}
M.~Najibi, P.~Samangouei, R.~Chellappa, and L.~S. Davis.
\newblock Ssh: Single stage headless face detector.
\newblock In {\em Proceedings of the IEEE Conference on Computer Vision and
  Pattern Recognition}, pages 4875--4884, 2017.

\bibitem{qin2016joint}
H.~Qin, J.~Yan, X.~Li, and X.~Hu.
\newblock Joint training of cascaded cnn for face detection.
\newblock In {\em Proceedings of the IEEE Conference on Computer Vision and
  Pattern Recognition}, pages 3456--3465, 2016.

\bibitem{ranjan2015deep}
R.~Ranjan, V.~M. Patel, and R.~Chellappa.
\newblock A deep pyramid deformable part model for face detection.
\newblock In {\em Biometrics Theory, Applications and Systems (BTAS), 2015 IEEE
  7th International Conference on}, pages 1--8. IEEE, 2015.

\bibitem{ranjan2017hyperface}
R.~Ranjan, V.~M. Patel, and R.~Chellappa.
\newblock Hyperface: A deep multi-task learning framework for face detection,
  landmark localization, pose estimation, and gender recognition.
\newblock {\em IEEE Transactions on Pattern Analysis and Machine Intelligence},
  2017.

\bibitem{redmon2017yolo9000}
J.~Redmon and A.~Farhadi.
\newblock Yolo9000: Better, faster, stronger.
\newblock In {\em Computer Vision and Pattern Recognition (CVPR), 2017 IEEE
  Conference on}, pages 6517--6525. IEEE, 2017.

\bibitem{ren2014face}
S.~Ren, X.~Cao, Y.~Wei, and J.~Sun.
\newblock Face alignment at 3000 fps via regressing local binary features.
\newblock In {\em Proceedings of the IEEE Conference on Computer Vision and
  Pattern Recognition}, pages 1685--1692, 2014.

\bibitem{ren2015faster}
S.~Ren, K.~He, R.~Girshick, and J.~Sun.
\newblock Faster r-cnn: Towards real-time object detection with region proposal
  networks.
\newblock In {\em Advances in neural information processing systems}, pages
  91--99, 2015.

\bibitem{ren2016single}
W.~Ren, S.~Liu, H.~Zhang, J.~Pan, X.~Cao, and M.-H. Yang.
\newblock Single image dehazing via multi-scale convolutional neural networks.
\newblock In {\em European conference on computer vision}, pages 154--169.
  Springer, 2016.

\bibitem{shrivastava2016training}
A.~Shrivastava, A.~Gupta, and R.~Girshick.
\newblock Training region-based object detectors with online hard example
  mining.
\newblock In {\em Proceedings of the IEEE Conference on Computer Vision and
  Pattern Recognition}, pages 761--769, 2016.

\bibitem{simonyan2014very}
K.~Simonyan and A.~Zisserman.
\newblock Very deep convolutional networks for large-scale image recognition.
\newblock {\em arXiv preprint arXiv:1409.1556}, 2014.

\bibitem{sindagi2017cnn}
V.~A. Sindagi and V.~M. Patel.
\newblock Cnn-based cascaded multi-task learning of high-level prior and
  density estimation for crowd counting.
\newblock In {\em Advanced Video and Signal Based Surveillance (AVSS), 2017
  14th IEEE International Conference on}, pages 1--6. IEEE, 2017.

\bibitem{sung1998example}
K.-K. Sung and T.~Poggio.
\newblock Example-based learning for view-based human face detection.
\newblock {\em IEEE Transactions on pattern analysis and machine intelligence},
  20(1):39--51, 1998.

\bibitem{taborsky2015annotating}
E.~Taborsky, K.~Allen, A.~Blanton, A.~K. Jain, and B.~F. Klare.
\newblock Annotating unconstrained face imagery: A scalable approach.
\newblock In {\em Biometrics (ICB), 2015 International Conference on}, pages
  264--271. IEEE, 2015.

\bibitem{viola2001rapid}
P.~Viola and M.~Jones.
\newblock Rapid object detection using a boosted cascade of simple features.
\newblock In {\em Computer Vision and Pattern Recognition, 2001. CVPR 2001.
  Proceedings of the 2001 IEEE Computer Society Conference on}, volume~1, pages
  I--I. IEEE, 2001.

\bibitem{viola2004robust}
P.~Viola and M.~J. Jones.
\newblock Robust real-time face detection.
\newblock {\em International journal of computer vision}, 57(2):137--154, 2004.

\bibitem{wang2018high}
L.~Wang, V.~Sindagi, and V.~Patel.
\newblock High-quality facial photo-sketch synthesis using multi-adversarial
  networks.
\newblock In {\em Automatic Face \& Gesture Recognition (FG 2018), 2018 13th
  IEEE International Conference on}, pages 83--90. IEEE, 2018.

\bibitem{xiong2013supervised}
X.~Xiong and F.~De~la Torre.
\newblock Supervised descent method and its applications to face alignment.
\newblock In {\em Computer Vision and Pattern Recognition (CVPR), 2013 IEEE
  Conference on}, pages 532--539. IEEE, 2013.

\bibitem{yan2014face}
J.~Yan, X.~Zhang, Z.~Lei, and S.~Z. Li.
\newblock Face detection by structural models.
\newblock {\em Image and Vision Computing}, 32(10):790--799, 2014.

\bibitem{yang2014aggregate}
B.~Yang, J.~Yan, Z.~Lei, and S.~Z. Li.
\newblock Aggregate channel features for multi-view face detection.
\newblock In {\em Biometrics (IJCB), 2014 IEEE International Joint Conference
  on}, pages 1--8. IEEE, 2014.

\bibitem{faceevaluation15}
B.~Yang, J.~Yan, Z.~Lei, and S.~Z. Li.
\newblock Fine-grained evaluation on face detection in the wild.
\newblock In {\em Automatic Face and Gesture Recognition (FG), 11th IEEE
  International Conference on}. IEEE, 2015.

\bibitem{yang2015facial}
S.~Yang, P.~Luo, C.-C. Loy, and X.~Tang.
\newblock From facial parts responses to face detection: A deep learning
  approach.
\newblock In {\em Proceedings of the IEEE International Conference on Computer
  Vision}, pages 3676--3684, 2015.

\bibitem{yang2016wider}
S.~Yang, P.~Luo, C.-C. Loy, and X.~Tang.
\newblock Wider face: A face detection benchmark.
\newblock In {\em Proceedings of the IEEE Conference on Computer Vision and
  Pattern Recognition}, pages 5525--5533, 2016.

\bibitem{yang2017face}
S.~Yang, Y.~Xiong, C.~C. Loy, and X.~Tang.
\newblock Face detection through scale-friendly deep convolutional networks.
\newblock {\em arXiv preprint arXiv:1706.02863}, 2017.

\bibitem{zhang2017rain}
H.~Zhang, V.~Sindagi, and V.~M. Patel.
\newblock Image de-raining using a conditional generative adversarial network.
\newblock {\em arXiv preprint arXiv:1701.05957}, 2017.

\bibitem{zhang2016joint}
K.~Zhang, Z.~Zhang, Z.~Li, and Y.~Qiao.
\newblock Joint face detection and alignment using multitask cascaded
  convolutional networks.
\newblock {\em IEEE Signal Processing Letters}, 23(10):1499--1503, 2016.

\bibitem{zhang2017s3fd}
S.~Zhang, X.~Zhu, Z.~Lei, H.~Shi, X.~Wang, and S.~Z. Li.
\newblock S3fd: Single shot scale-invariant face detector.
\newblock In {\em Proceedings of the IEEE Conference on Computer Vision and
  Pattern Recognition}, pages 192--201, 2017.

\bibitem{zhang2017s}
S.~Zhang, X.~Zhu, Z.~Lei, H.~Shi, X.~Wang, and S.~Z. Li.
\newblock S3fd: Single shot scale-invariant face detector.
\newblock In {\em Proceedings of the IEEE Conference on Computer Vision and
  Pattern Recognition}, 2017.

\bibitem{zhu2017cms}
C.~Zhu, Y.~Zheng, K.~Luu, and M.~Savvides.
\newblock Cms-rcnn: contextual multi-scale region-based cnn for unconstrained
  face detection.
\newblock In {\em Deep Learning for Biometrics}, pages 57--79. Springer, 2017.

\bibitem{zhu2012face}
X.~Zhu and D.~Ramanan.
\newblock Face detection, pose estimation, and landmark localization in the
  wild.
\newblock In {\em Computer Vision and Pattern Recognition (CVPR), 2012 IEEE
  Conference on}, pages 2879--2886. IEEE, 2012.

\end{thebibliography}
}

\end{document}